\documentclass{article}

\usepackage[final]{neurips_data_2022}

\usepackage[utf8]{inputenc} % allow utf-8 input
\usepackage[T1]{fontenc}    % use 8-bit T1 fonts
\usepackage[colorlinks=true,urlcolor=MidnightBlue,citecolor=MidnightBlue,linkcolor=MidnightBlue,backref=page]{hyperref}       % hyperlinks
\usepackage{url}            % simple URL typesetting
\usepackage{booktabs}       % professional-quality tables
\usepackage{amsfonts}       % blackboard math symbols
\usepackage{nicefrac}       % compact symbols for 1/2, etc.
\usepackage{microtype}      % microtypography
\usepackage{graphicx}
\usepackage[dvipsnames]{xcolor}
\usepackage{xspace}
\usepackage{amsmath}
\usepackage[capitalise]{cleveref}
\usepackage{subcaption}
\usepackage{siunitx}[=v2]
\usepackage{dirtree}
\usepackage[inline]{enumitem}
\usepackage{microtype}
\usepackage{inconsolata}

\makeatletter
\newcommand{\subalign}[1]{%
  \vcenter{%
    \Let@ \restore@math@cr \default@tag
    \baselineskip\fontdimen10 \scriptfont\tw@
    \advance\baselineskip\fontdimen12 \scriptfont\tw@
    \lineskip\thr@@\fontdimen8 \scriptfont\thr@@
    \lineskiplimit\lineskip
    \ialign{\hfil$\m@th\scriptstyle##$&$\m@th\scriptstyle{}##$\hfil\crcr
      #1\crcr
    }%
  }%
}
\makeatother

\def\dataset{MoCapAct\xspace}
\def\datasetfull{Motion Capture with Actions\xspace}

\title{\dataset: A Multi-Task Dataset for \\ Simulated Humanoid Control}

\def\dmcontrol{\texttt{dm\char`_control}\xspace}
\def\encoder{\mathrm{enc}}
\def\decoder{\mathrm{dec}}
\def\reference{\mathrm{ref}}
\def\mocap{MoCap\xspace}
\def\projectwebsite{\href{https://microsoft.github.io/MoCapAct}{project website}\xspace}

\let\originalleft\left
\let\originalright\right
\renewcommand{\left}{\mathopen{}\mathclose\bgroup\originalleft}
\renewcommand{\right}{\aftergroup\egroup\originalright}

\author{%
  Nolan Wagener\thanks{Correspondence to \href{mailto:nolan.wagener@gatech.edu}{\texttt{nolan.wagener@gatech.edu}} and \href{mailto:matthew.hausknecht@gmail.com}{\texttt{matthew.hausknecht@gmail.com}}}~~$^{\!1}$ \quad Andrey Kolobov$^2$ \quad Felipe Vieira Frujeri$^2$ \\ \textbf{Ricky Loynd}$^2$ \quad \textbf{Ching-An Cheng}$^2$ \quad \textbf{Matthew Hausknecht}$^{*2}$\\
  $^1$Institute for Robotics and Intelligent Machines, Georgia Institute of Technology \\
  $^2$Microsoft Research
 }

\begin{document}

\maketitle

% Reset footnote counter since \thanks counts as a footnote
\setcounter{footnote}{0}

\begin{abstract}
Simulated humanoids are an appealing research domain due to their physical capabilities.
Nonetheless, they are also challenging to control, as a policy must drive an unstable, discontinuous, and high-dimensional physical system.
One widely studied approach is to utilize motion capture (\mocap) data to teach the humanoid agent low-level skills (e.g., standing, walking, and running) that can then be re-used to synthesize high-level behaviors.
However, even with \mocap data, controlling simulated humanoids remains very hard, as \mocap data offers only kinematic information. Finding physical control inputs to realize the demonstrated motions requires computationally intensive methods like reinforcement learning.
Thus, despite the publicly available \mocap data, its utility has been limited to institutions with large-scale compute.
In this work, we dramatically lower the barrier for productive research on this topic by training and releasing high-quality agents that can track over three hours of \mocap data for a simulated humanoid in the \dmcontrol physics-based environment.
We release \emph{\mbox{\dataset}}~(\datasetfull), a dataset of these expert agents and their rollouts, which contain proprioceptive observations and actions.
We demonstrate the utility of \mbox{\dataset} by using it to train a \emph{single} hierarchical policy capable of tracking the \emph{entire} \mocap dataset within \dmcontrol and show the learned low-level component can be re-used to efficiently learn downstream high-level tasks.
Finally, we use \mbox{\dataset} to train an autoregressive GPT model and show that it can control a simulated humanoid to perform natural motion completion given a motion prompt.
Videos of the results and links to the code and dataset are available at the \projectwebsite.

\end{abstract}

\section{Introduction}

The wide range of human physical capabilities makes simulated humanoids a compelling platform for studying motor intelligence.
Learning and utilization of motor skills is a prominent research topic in machine learning, with advances ranging from emergence of learned locomotion skills in traversing an obstacle course~\citep{heess2017emergence} to the picking up and carrying of objects to desired locations~\citep{merel2020catch,peng2019mcp} to team coordination in simulated soccer~\citep{liu2022motor}.
Producing natural and physically plausible human motion animation~\citep{harvey2020robust,kania2021trajevae,yuan2020residual} is an active research topic in the game and movie industries.
However, while physical simulation of human capabilities is a useful research domain, it is also very challenging from a control perspective.
A controller must contend with an unstable, discontinuous, and high-dimensional system that requires a high degree of coordination to execute a desired motion.

\textit{Tabula rasa} learning of complex humanoid behaviors (e.g., navigating through an obstacle field) is extremely difficult for all known learning approaches.
In light of this challenge, motion capture~(\mocap) data has become an increasingly common aid in humanoid control research~\citep{merel2017learning,peng2018deepmimic}.
\mocap trajectories contain kinematic information about motion: they are sequences of configurations and poses that the human body assumes throughout the motion in question.
This data can alleviate the difficulty of training sophisticated control policies by enabling a simulated humanoid to learn \emph{low-level} motor skills from \mocap demonstrations.
The low-level skills can then be re-used for learning advanced, higher-level motions.
Datasets such as CMU \mocap~\citep{cmu2003mocap}, Human3.6M~\citep{ionescu2013human3}, and LaFAN1~\citep{harvey2020robust} offer hours of recorded human motion, ranging from simple locomotion demonstrations to interactions with other humans and objects.

However, since \mocap data only offers kinematic information, utilizing it in a physics simulator requires recovering the actions (e.g., joint torques) that induce the sequence of kinematic poses in a given \mocap trajectory (i.e., track the trajectory).
While easier than \textit{tabula rasa} learning of a high-level task, finding an action sequence that makes a humanoid track a \mocap sequence is still non-trivial.
For instance, this problem has been tackled with reinforcement learning~\citep{chentanez2018physics,merel2019neural,peng2018deepmimic} and adversarial learning~\citep{merel2017learning,wang2017robust}.
The computational burden of finding these actions scales with the amount of \mocap data, and training agents to recreate hours of \mocap data requires significant compute. \emph{As a result, despite the broad availability of \mocap datasets, their utility---and their potential for enabling research progress on learning-based humanoid control---has been limited to institutions with large compute budgets.} 
\begin{figure}[t]
    \centering
    \includegraphics[width=\textwidth]{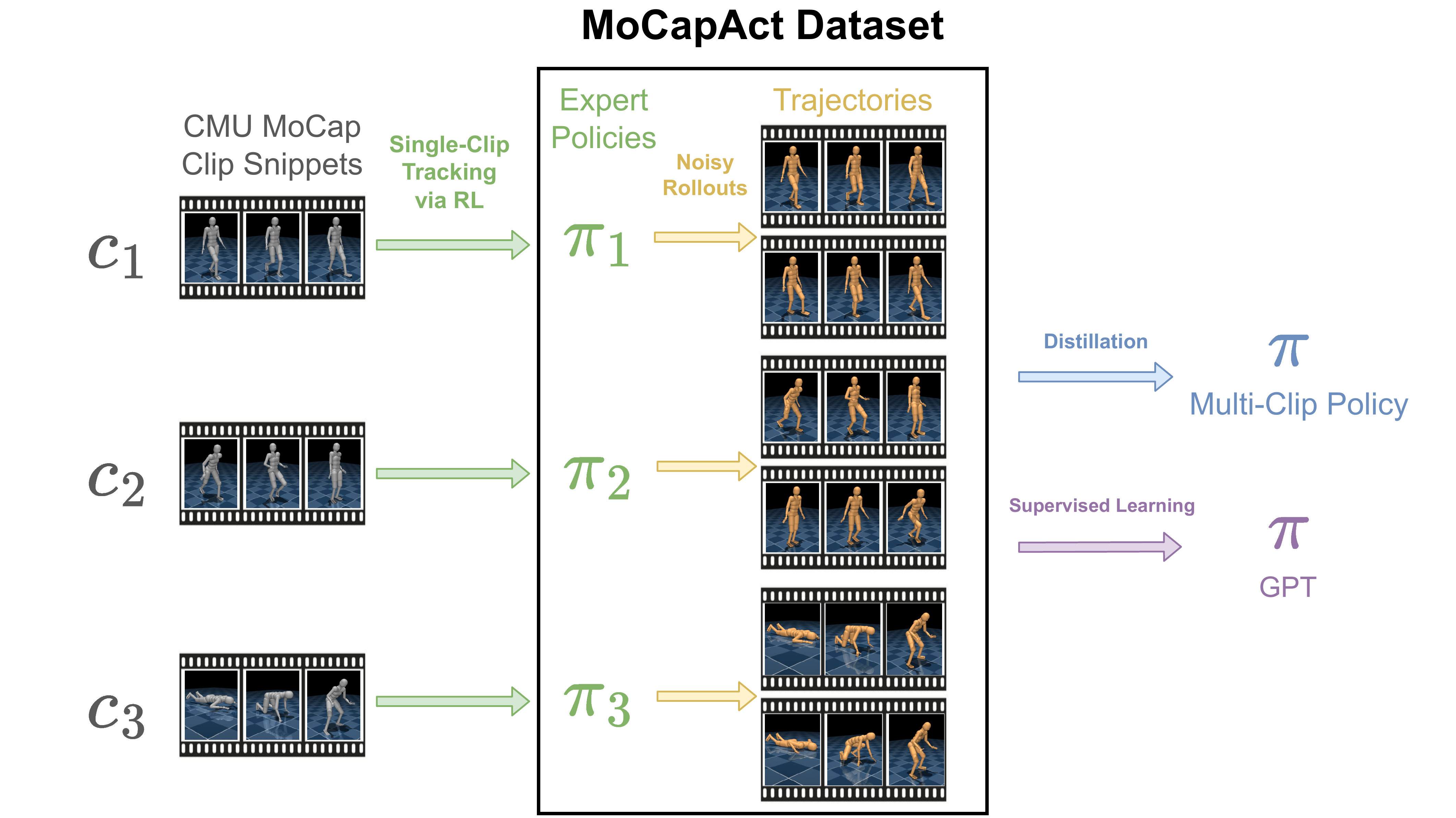}
    \caption{The \textbf{\dataset Dataset} includes expert policies that are trained to track individual clips. A dataset of noise-injected rollouts (containing observations and actions) is then collected from each expert. These rollouts can subsequently be used to, for instance, train a multi-clip or GPT policy.}
    \label{fig:MoCapAct}
\end{figure}

To remove this obstacle and facilitate the use of \mocap data in humanoid control research, we introduce \textbf{\dataset}~(\datasetfull, \cref{fig:MoCapAct}), a dataset of high-quality \mocap-tracking policies for a MuJoCo-based~\citep{todorov2012mujoco} simulated humanoid as well as a collection of rollouts from these expert policies. The policies from \dataset\ can track 3.5 hours of recorded motion from CMU \mocap~\citep{cmu2003mocap}, one of the largest publicly available \mocap datasets.
We analyze the expert policies of \dataset and, to illustrate \dataset's usefulness for learning diverse motions, use the expert rollouts to train a \textit{single} hierarchical policy which is capable of tracking \emph{all} of the considered \mocap clips.
We then re-use the low-level component of the policy to efficiently learn downstream tasks via reinforcement learning.
Finally, we use the dataset for generative motion completion by training a GPT network~\citep{kaparthy2020mingpt} to produce a motion in the MuJoCo simulator given a motion prompt.

\section{Related Work} \label{sec:related}

\paragraph{\mocap Data}
Of the existing datasets featuring motion capture of humans, the largest and most cited are CMU \mocap~\citep{cmu2003mocap} and Human3.6M~\citep{ionescu2013human3}.
These datasets feature tens of hours of human motion capture arranged as a collection of clips recorded at 30-120Hz. They demonstrate a wide range of motions, including locomotion (e.g., walking, running, jumping, and turning), physical activities (e.g., dancing, boxing, and gymnastics), and interactions with other humans and objects.

\paragraph{\mocap Tracking via Reinforcement Learning}
To make use of \mocap data for downstream tasks, much of prior work first learns individual clip-tracking policies.
\citet{peng2018deepmimic} and \citet{merel2019hierarchical,merel2019neural,merel2020catch} use reinforcement learning (RL) to learn the clip-tracking policies, whereas \citet{merel2017learning} use adversarial imitation learning.
Upon learning the tracking policies, there are a variety of ways to utilize them.
\citet{peng2018deepmimic} and \citet{merel2017learning,merel2019hierarchical} learn a skill-selecting policy to dynamically choose a clip-tracking policy to achieve new tasks.
\citet{merel2019neural,merel2020catch} instead opt for a distillation approach, whereby they collect rollouts from the clip-tracking policies and then train a hierarchical multi-clip policy via supervised learning on the rollouts.
The low-level policy is then re-used to aid in learning new high-level tasks.

Alternatively, large-scale RL may be used to learn a single policy that covers the \mocap dataset.
\citet{hasenclever2020comic} use a distributed RL setup for the MuJoCo simulator~\citep{todorov2012mujoco}, while \citet{peng2022ase} use the GPU-based Isaac simulator~\citep{makoviychuk2021isaac} to perform RL on a single machine.

While some prior work has released source code to train individual clip-tracking policies~\citep{peng2018deepmimic,yuan2020residual}, their included catalog of policies is small, and the resources needed to train per-clip policies scale linearly with the number of \mocap clips.
In the process of our work, we found that we needed about \emph{50 years} of wall-clock time to train the policies to track our \mocap corpus using a similar approach to \citet{peng2018deepmimic}.

\paragraph{Motion Completion} Outside of the constraints of a physics simulator, learning natural completions of \mocap trajectories (i.e., producing a trajectory given a prompt trajectory) is the subject of many research papers~\citep{mourot2022survey}, typically motivated by the challenging and labor-intensive process of creating realistic animations for video games and films.
Prior work
~\citep{aksan2021spatio,harvey2020robust,kania2021trajevae,mao2019learning,tevet2022motionclip,wang2019imitation}
typically trains a model to replicate the kinematic motion found in a \mocap dataset, which is then evaluated according to how well the model can predict or synthesize motions given some initial prompt on held-out trajectories.

The more difficult task of performing motion completion \emph{within a physics simulator} is not widely studied.
\citet{yuan2020residual} jointly learn a kinematic policy and a tracking policy, where the kinematic policy predicts future kinematic poses given a recent history of observations and the tracking policy outputs a low-level action to track the predicted poses.

\section{The \dmcontrol Humanoid Environment} \label{sec:environment}

\begin{figure}[t]
    \centering
    \includegraphics[width=0.16\textwidth]{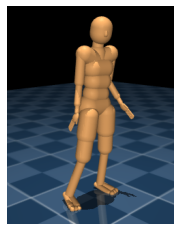} \hfill
    \includegraphics[width=0.16\textwidth]{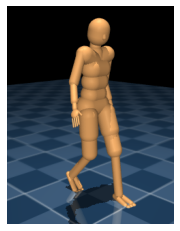} \hfill
    \includegraphics[width=0.16\textwidth]{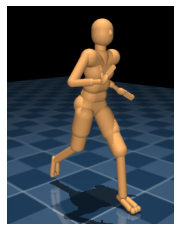} \hfill
    \includegraphics[width=0.16\textwidth]{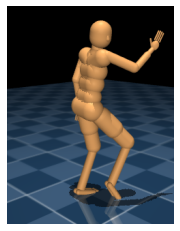} \hfill
    \includegraphics[width=0.16\textwidth]{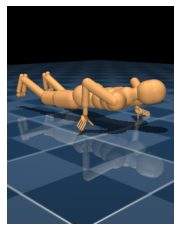}
    \caption{The humanoid displaying a variety of motions from the CMU \mocap dataset.}
    \label{fig:cmu humanoid}
\end{figure}

Our simulated humanoid of interest is the ``CMU Humanoid'' (\cref{fig:cmu humanoid}) from the \href{https://github.com/deepmind/dm_control}{\dmcontrol package}~\citep{tunyasuvunakool2020dm_control}, which contains 56 joints and is designed to be similar to an average human body.
The humanoid contains a rich and customizable observation space, from proprioceptive observations like joint positions and velocities, actuator states, and touch sensor measurements to high-dimensional observations like images from an egocentric camera.
The action~$a$ is the desired joint angles of the humanoid, which are then converted to joint torques via some pre-defined PD controllers.
The humanoid operates in the MuJoCo simulator~\citep{todorov2012mujoco}.

The \dmcontrol package contains a variety of tools for the humanoid.
The package comes with pre-defined tasks like navigation through an obstacle field~\citep{heess2017emergence}, maze navigation~\citep{merel2019hierarchical}, and soccer~\citep{liu2022motor}, and a user may create custom tasks with the package's API.
The \dmcontrol package also integrates 3.5 hours of motion sequences from the \href{http://mocap.cs.cmu.edu/}{CMU Motion Capture Dataset}~\citep{cmu2003mocap}, including clips of locomotion (standing, walking, turning, running, jumping, etc.), acrobatics, and arm movements.
Each clip~$C$ is a reference state sequence~$(\hat{s}^C_0, \hat{s}^C_1, \dots, \hat{s}^C_{T_C-1})$, where $T_C$ is the clip length and each $\hat{s}^C_t$ contains kinematic information like joint angles, joint velocities, and humanoid pose.

As discussed in \cref{sec:related}, training a control policy to work on all of the included clips requires large-scale solutions.
For example, \citet{hasenclever2020comic} rely on a distributed RL approach that uses about ten billion environment interactions collected by 4000 parallel actor processes running for multiple days.
To our knowledge, there are no agents publicly available that can track all the \mocap data within \dmcontrol.
We address this gap by releasing a dataset of high-quality experts and their rollouts for the ``CMU Humanoid'' in the \dmcontrol package.

\section{\dataset Dataset}

The \dataset dataset~(\cref{fig:MoCapAct}) consists of:
\begin{itemize}
\item experts each trained to track an individual snippet from the \mocap dataset (\cref{sec:clip experts}) and
\item HDF5 files containing rollouts from the aforementioned experts (\cref{sec:expert rollouts}).
\end{itemize}

We include documentation of the \dataset dataset in \cref{app:dataset}.

\subsection{Clip Snippet Experts} \label{sec:clip experts}

Our expert training scheme largely follows that of \citet{merel2019hierarchical,merel2019neural} and \citet{peng2018deepmimic}, which we now summarize.

\paragraph{Training}
We split each clip in the MoCap dataset into 4--6 second snippets with 1-second overlaps.
With 836 clips in the \mocap dataset, this clip splitting results in 2589 snippets.
For each clip snippet $c$, we train a time-indexed Gaussian policy $\pi_c(a|s,t)$ to track the snippet.
We use the same clip-tracking reward function $r_c(s,t)$ as \citet{hasenclever2020comic}, which encourages matching the MoCap clip's joint angles and velocities, positions of various body parts, and joint orientations.
This reward function lies in the interval $[0, 1.4]$.
To speed up training, we use the same early episode termination condition as \citet{hasenclever2020comic}, which activates if the humanoid deviates too far from the snippet.
To help exploration, the initial state of an episode is generated by randomly sampling a time step from the given snippet.
The Gaussian policy $\pi_c$ uses a mean parameterized by a neural network as well as a fixed standard deviation of $0.1$ for each action to induce robustness and to prepare for the noisy rollouts (\cref{sec:expert rollouts}).
We use the Stable-Baselines3~\citep{raffin2021stable} implementation of PPO~\citep{schulman2017proximal} to train the experts.
Our training took about \emph{50 years} of wall-clock time.
We give hyperparameters and training details in \cref{app:clip experts}.

\paragraph{Results}
To account for the snippets having different lengths and for the episode initialization scheme used in training, we report our evaluations in a length-normalized fashion.\footnote{We point out that PPO uses the original unnormalized reward for policy optimization.}
For a snippet~$c$ (with length~$T_c$) and some policy~$\pi$, recall that we initialize the humanoid at some randomly chosen time step $t_0$ from $c$ and then generate the trajectory~$\tau$ by rolling out $\pi$ from $t_0$ until either the end of the snippet or early termination.
Let $R(\tau)$ and $L(\tau)$ denote the accumulated reward and the length of the trajectory~$\tau$, respectively.
We define the normalized episode reward and normalized episode length of $\tau$ as $\frac{R(\tau)}{T_c - t_0}$ and $\frac{L(\tau)}{T_c - t_0}$, respectively.
One consequence of this definition is that trajectories that are terminated early in a snippet yield smaller normalized episode rewards and lengths.
Next, we define the average normalized episode reward and average normalized episode length of policy $\pi$ on snippet $c$ as $\hat{R}_c(\pi) = \mathbb{E}_{t_0 \sim c} \mathbb{E}_{\tau \sim \pi | t_0}\left[ \frac{R(\tau)}{T_c - t_0} \right]$ and $\hat{L}_c(\pi) = \mathbb{E}_{t_0 \sim c} \mathbb{E}_{\tau \sim \pi | t_0}\left[ \frac{L(\tau)}{T_c - t_0} \right]$, respectively.
For example, if $\pi$ always successfully tracks some \mocap snippet from any $t_0$ to the end of the snippet, $\pi$ has an average normalized episode length of $1$ on snippet $c$.

\begin{table}[t!]
    \centering
    \caption{Snippet expert results on the \mocap snippets within \dmcontrol.
    We disable the Gaussian noise for $\pi_c$ when computing these results.}
    \begin{tabular}{r|c|c|c|c|c}
                              & {\small Mean}    & {\small Standard deviation} & {\small Median} & {\small Minimum} & {\small Maximum} \\ \hline
    {\small Average normalized episode reward} & $0.816$ & $0.153$            & $0.777$ & $0.217$ & $1.233$ \\
    {\small Average normalized episode length} & $0.997$ & $0.022$            & $1.000$ & $0.424$ & $1.000$
    \end{tabular}
    \label{tab:clip expert metrics}
\end{table}
\begin{figure}[t!]
    \centering
    \includegraphics[width=0.9\textwidth]{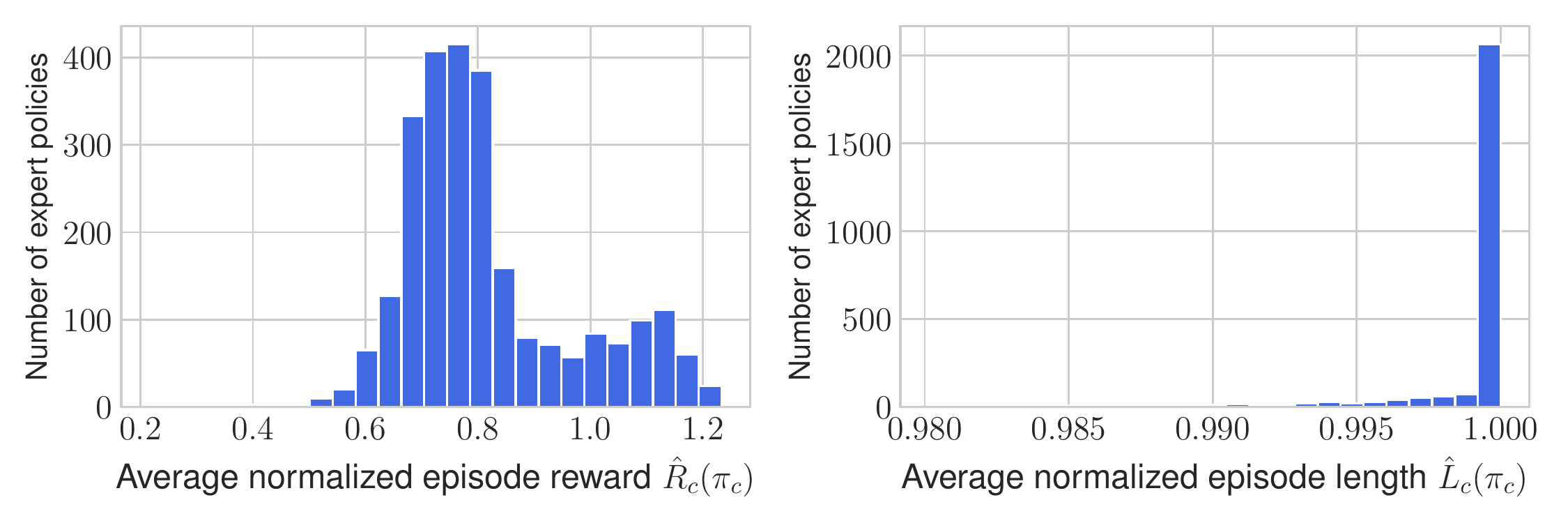}
    \caption{Clip expert results on the \mocap snippets within \dmcontrol.} 
    \label{fig:clip expert metrics}
\end{figure}

Overall, the clip experts reliably track the overwhelming majority of the MoCap snippets (\cref{tab:clip expert metrics} and \cref{fig:clip expert metrics}).
Averaged over all the snippets, the experts have a per-joint mean angle error of $0.062$ radians.
We find that $80\%$ of the trained experts have an average normalized episode length of at least~$0.999$.
We also observe there is a bimodal structure to the reward distribution in \cref{fig:clip expert metrics}, which is due to many clips having artifacts like jittery limbs and extremities clipping through the ground.
These artifacts limit the extent to which the humanoid can track the clip.
Among the handful of experts with very low reward (between $0.2$ and $0.5$), we find that the corresponding clips are erroneously sped up, making them impossible to track in the simulator.

\begin{figure}[b]
    \centering
    \includegraphics[width=0.9\textwidth]{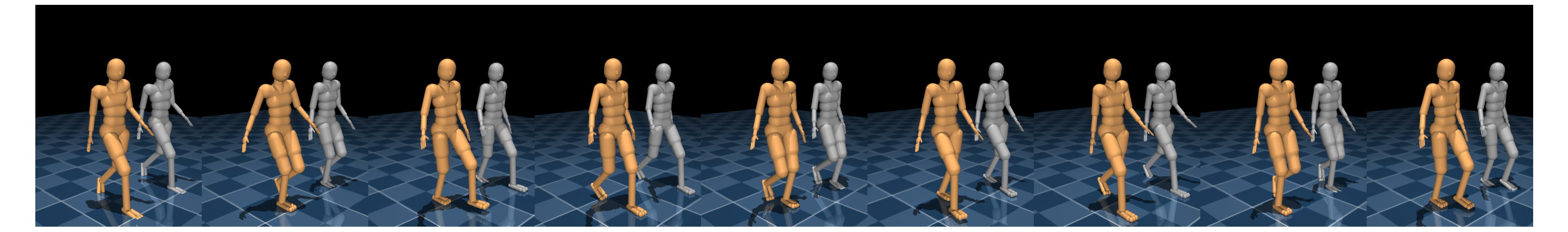} \\
    \includegraphics[width=0.9\textwidth]{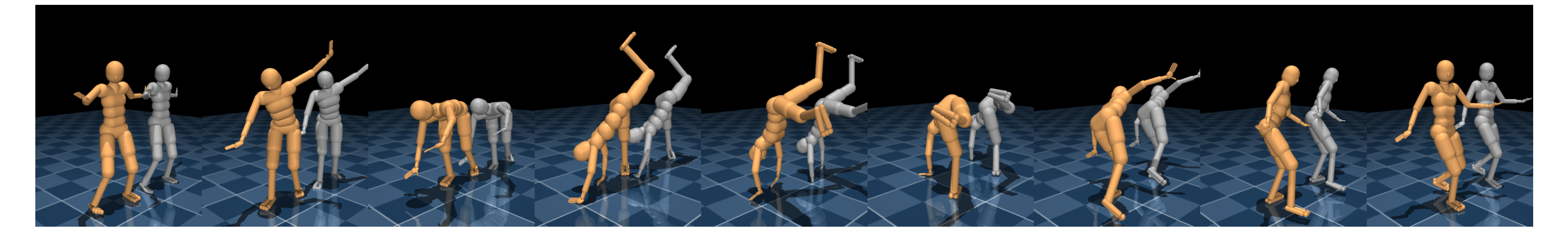} \\
    \includegraphics[width=0.9\textwidth]{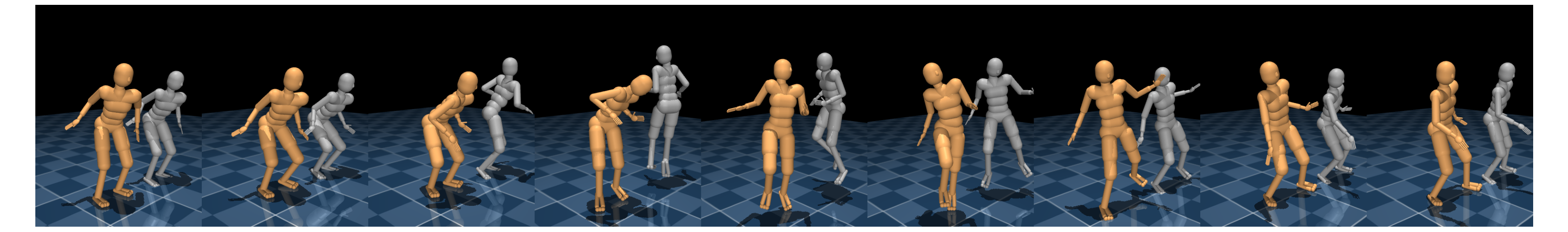}
    \caption{Visualizations of clip experts.
    The top two rows show episodes (first: walking, second:~cartwheel) where the expert (bronze humanoid) closely tracks the corresponding \mocap clip (grey humanoid).
    The bottom row shows a clip where the expert and \mocap clip differ in behavior.
    The \mocap clip demonstrates a 360-degree jump, whereas the expert jumps without spinning.}
    \label{fig:expert montage}
\end{figure}

The experts produce motion that is generally indistinguishable from the \mocap reference (\cref{fig:expert montage}), from simple walking behaviors seen in the top row to highly coordinated motions like the cartwheel in the middle row.
On some clips, the expert deviates from the clip because the demonstrated motion is too highly dynamic, such as the 360-degree jump in the bottom row.
Instead, the expert typically learns some other behavior that keeps the episode from terminating early, which in this case is jumping without spinning.
We also point out that, in these failure modes, the humanoid still tracks some portions of the reference, such as hand positions and orientations.
\citet{yuan2020residual} rectify similar tracking issues by augmenting the action space with external forces on certain parts of the humanoid body, but we do not explore this avenue since the issue only affects a small number of clips.
We encourage the reader to visit the \projectwebsite to see  videos of the clip experts.

\subsection{Expert Rollouts} \label{sec:expert rollouts}

Following \citet{merel2019neural}, we roll out the experts on their respective snippets and collect data from the rollouts into a dataset~$\mathcal{D}$.
In order to obtain a broad state coverage from the experts, we repeatedly roll out the \emph{stochastic} experts (i.e., with Gaussian noise injected into the actions) starting from different initial states.
This injected noise helps the dataset cover states that a policy learned by imitating the dataset would visit, therefore mitigating the distribution shift issue for the learned policy~\citep{laskey2017dart,merel2019neural}. 

For each clip snippet~$c$, we denote the corresponding expert policy as ${\pi_c(a|s,t) = \mathcal{N}(a; \mu_c(s,t), 0.1^2 I)}$, where $\mu_c(s,t)$ is the mean of the expert's action distribution.
We initialize the humanoid at some point in the snippet~$c$ (half of the time at the beginning of the snippet and otherwise at some random point in the snippet).
We then roll out $\pi_c$ until either the end of the snippet or early termination using the scheme from \cref{sec:clip experts}.
At every time step~$t$ in the rollout, we log the humanoid state~$s_t$, the target reference poses~$s_t^\reference = (\hat{s}_{t+1}^c, \dots, \hat{s}_{t+5}^c)$ from the next five steps of the \mocap snippet, the expert's sampled action~$a_t$, the expert's mean action~$\bar{a}_t = \mu_c(s_t,t)$, the observed snippet reward~$r_c(s_t,t)$, the estimated value~$\hat{V}^{\pi_c}(s_t)$, and the estimated advantage~$\hat{A}^{\pi_c}(s_t,a_t)$ into HDF5 files.

We release two versions of the rollout dataset:
\begin{itemize}
\item a ``large'' $600$-gigabyte collection at 200 rollouts per snippet with a total of 67 million environment transitions (corresponding to $620$ hours in the simulator) and
\item a ``small'' $50$-gigabyte collection at 20 rollouts per snippet with a total of 5.5 million environment transitions (corresponding to $51$ hours in the simulator).
\end{itemize}
In our application of \dataset~(\cref{sec:applications}), we use the ``large'' version of the dataset.
We do observe, though, that the multi-clip policy results~(\cref{sec:npmp}) are similar when using either dataset.

\section{Applications} \label{sec:applications}

\begin{figure}[t]
    \centering
    \begin{subfigure}[b]{0.66\textwidth}
        \centering
        \includegraphics[width=\textwidth]{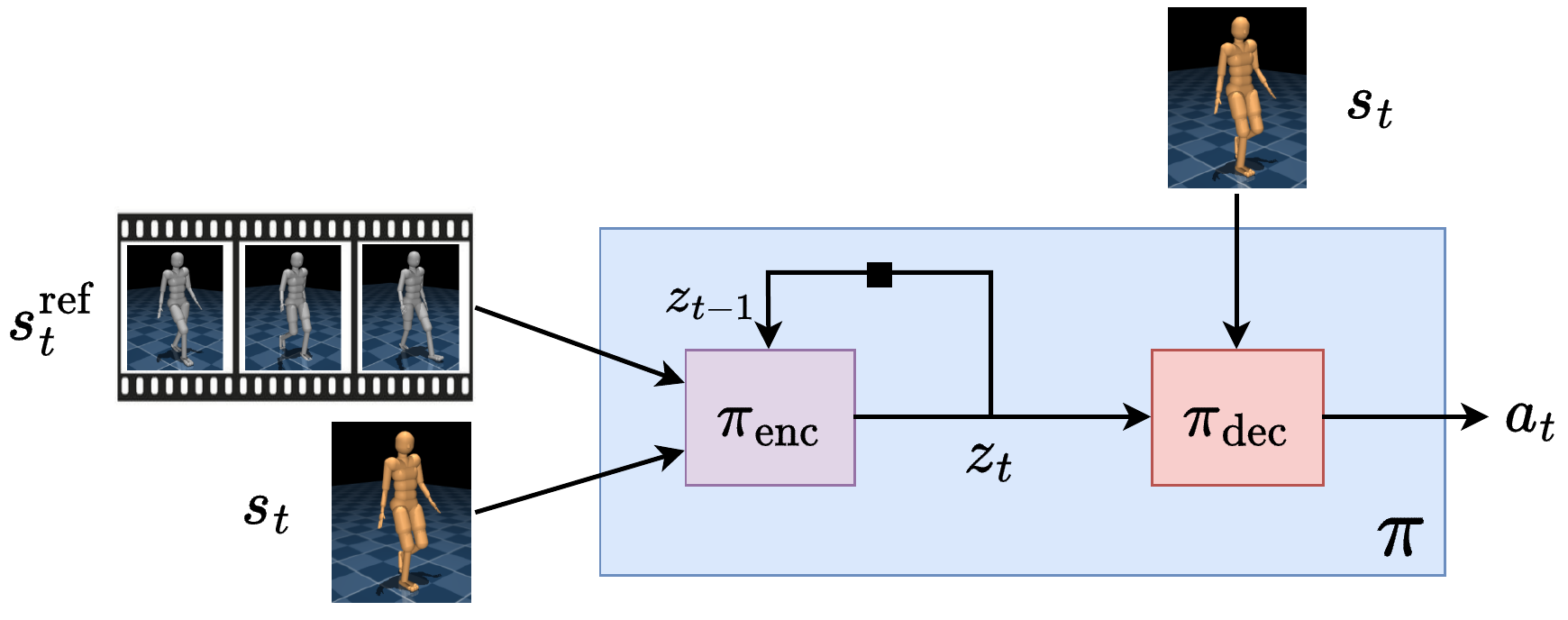}
        \caption{Multi-clip tracking policy.}
        \label{fig:multiclip policy}
    \end{subfigure}
    \hfill
    \begin{subfigure}[b]{0.29\textwidth}
        \centering
        \includegraphics[width=\textwidth]{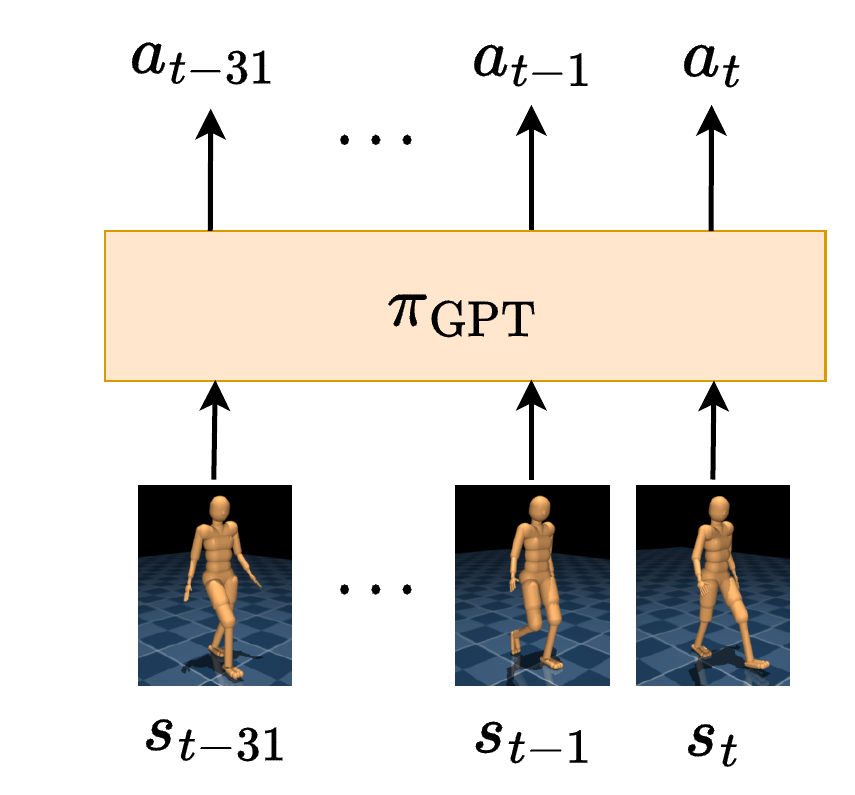}
        \caption{GPT policy.}
        \label{fig:gpt}
    \end{subfigure}
    \caption{Policies used in the applications.}
    \label{fig:application policies}
\end{figure}

We train two policies~(\cref{fig:application policies}) using our dataset:
\begin{enumerate}
\item A hierarchical policy which can track all the \mocap snippets and be re-used for learning new high-level tasks (\cref{sec:npmp}).
\item An autoregressive GPT model which generates motion from a given prompt (\cref{sec:gpt}).
\end{enumerate}

\subsection{Multi-Clip Tracking Policy} \label{sec:npmp}

We first show the \dataset dataset can reproduce the results in \citet{merel2019neural} by learning a single policy that tracks the entire \mocap dataset within \dmcontrol.
Our policy architecture~(\cref{fig:multiclip policy}) follows the same encoder-decoder scheme as \citet{merel2019neural}, who introduce a ``motor intention''~$z_t$ which acts as a low-dimensional embedding of the \mocap reference~$s_t^\reference$.
The intention~$z_t$ is then decoded into an action~$a_t$.
In other words, the policy~$\pi$ is factored into an encoder~$\pi_\encoder$ and a decoder~$\pi_\decoder$.
The encoder~$\pi_\encoder(z_t|s_t, s_t^\reference, z_{t-1})$ compresses the \mocap reference~$s_t^\reference$ into an intention~$z_t$ and may use the current humanoid state~$s_t$ and previous intention~$z_{t-1}$ in predicting the current intention.
Furthermore, the encoder outputs an intention which is \emph{stochastic}, which models ambiguity in the \mocap reference and allows for the high-level behavior to be specified more coarsely.
The decoder~$\pi_\decoder(a_t | s_t, z_t)$ translates the sampled intention~$z_t$ into an action~$a_t$ with the aid of the state~$s_t$ as an additional input.

\subsubsection{Training} \label{sec:multi-clip policy training}
In our implementation, the encoder outputs the mean and diagonal covariance of a Gaussian distribution over a 60-dimensional motor intention $z_t$.
The decoder outputs the mean of a Gaussian distribution over actions with a standard deviation of $0.1$ for each action.
In training, we maximize a variant of the multi-step imitation learning objective from \citet{merel2019neural}:%
\begin{small}
\[
\mathbb{E}_{\subalign{(s_{1:T}, s_{1:T}^\reference, \bar{a}_{1:T}, c) &\sim \mathcal{D}, \\ z_{0:T} &\sim \pi_\encoder}} \left[ \sum_{t=1}^T \left[ w_c(s_t, \bar{a}_t) \log \pi_\decoder(\bar{a}_t | s_t, z_t) - \beta \, \mathrm{KL}(\pi_\encoder (z_t | s_t, s_t^\reference, z_{t-1}) \,\|\, p(z_t | z_{t-1})) \right] \right],
\]
\end{small}%
where $T$ is the sequence length, $w_c$ is a clip-dependent data-weighting function, $p(z_t | z_{t-1})$ is an autoregressive prior, and $\beta$ is a hyperparameter.

The weighting function $w_c$ allows for some data points to be considered more heavily, which may be useful given the spectrum of expert performance.
Letting $\lambda$ be a hyperparameter, we consider the following four weighting schemes:
\begin{itemize}
\item Behavioral cloning~(BC): $w_c(s, a) = 1$.
This scheme is commonly used in imitation learning and treats every data point equally.
\item Clip-weighted regression~(CWR): $w_c(s, a) = \exp( \hat{R}_c(\pi_c) / \lambda )$.
This scheme upweights data from snippets where the experts have higher average normalized rewards.
\item Advantage-weighted regression~(AWR)~\citep{peng2019advantage}: $w_c(s, a) = \exp( \hat{A}^{\pi_c}(s, a) / \lambda )$.
This scheme upweights actions that perform better than the expert's average return.
\item Reward-weighted regression~(RWR)~\citep{peters2007reinforcement}: \\ ${w_c(s, a) = \exp( \hat{Q}^{\pi_c}(s, a) / \lambda)}$, where $\hat{Q}^{\pi_c}(s, a) = \hat{V}^{\pi_c}(s) + \hat{A}^{\pi_c}(s, a)$.
This scheme upweights state-actions which have higher returns, which typically happens with good experts at earlier time steps in the corresponding snippet.
\end{itemize}

The KL divergence term encourages the decoder to follow a simple random walk.
In this case, the prior has the form $p(z_t | z_{t-1}) = \mathcal{N}(z_t; \alpha z_{t-1}, \sigma^2 I)$, where $\alpha \in [0, 1]$ is a hyperparameter and ${\sigma = \sqrt{1-\alpha^2}}$.
This prior in turn encourages the marginals to be a spherical Gaussian, i.e., ${p(z_t) = \mathcal{N}(z_t; 0, I)}$.
Furthermore, the regularization introduces a bottleneck~\citep{alemi2017deep} that limits the information the intention $z_t$ can provide about the state $s_t$ and \mocap reference $s_t^\reference$.
This forces the encoder to only encode high-level information about the reference (e.g., direction of motion of leg) while excluding fine-grained details (e.g., precise velocity of each joint in leg).

In our experiments, we found that the training takes about three hours on a single-GPU machine.
More training details are available in \cref{app:distillation}.

\paragraph{Results}
All four regression approaches yield broadly good results (\cref{tab:distillation metrics}), achieving $80\%$ to $84\%$ of the experts' performance on the \mocap dataset (cf. \cref{tab:clip expert metrics}).
We also see that every weighted regression scheme gives some improvement over the unweighted approach.
AWR only gives $1\%$ improvement over BC, likely because the experts are already near-optimal and the dataset lacks sufficient state-action coverage to reliably contain advantageous actions.
CWR gives a $3\%$ improvement over BC, which arises from the objective placing more emphasis on data coming from high-reward clips.
Finally, RWR gives a $5\%$ improvement over BC, which comes from increased weight on earlier time steps in high-reward clips.
This is a sensible weighting scheme since executing a skill requires taking correct actions at earlier time steps before completing the skill at later time steps.
As a point of comparison to prior work, the RWR-trained policy achieves an average reward-per-step (i.e., $\mathbb{E}[ R(\tau)/L(\tau) ]$) of $0.67$ on the ``Locomotion'' subset of the \mocap data, which is $96\%$ of the reward-per-step achieved by the large-scale RL approach of \citet{hasenclever2020comic}.
We also find that the RWR-trained policy has a per-joint mean angle error of $0.085$ radians.

\begin{table}[t]
    \footnotesize
    \centering
    \caption{Multi-clip results on the \mocap snippets, showing the mean and standard deviation over three seeds.
    For evaluation, we disable the Gaussian noise for $\pi_\decoder$ but keep the stochasticity for $\pi_\encoder$.}
    \begin{tabular}{r||c|c|c|c}
    & BC    & CWR & AWR & RWR \\ \hline
    Avg. normalized episode reward & $0.654 \pm 0.005$ & $0.671 \pm 0.003$ & $0.661 \pm 0.003$ & $\mathbf{0.688 \boldsymbol{\pm} 0.002}$ \\
    Avg. normalized episode length & $0.855 \pm 0.004$ & $0.858 \pm 0.003$ & $0.861 \pm 0.001$ & $\mathbf{0.868 \boldsymbol{\pm} 0.002}$
    \end{tabular}
    \label{tab:distillation metrics}
\end{table}

\begin{figure}[t]
    \centering
    \begin{subfigure}{0.45\textwidth}
        \includegraphics[width=\linewidth]{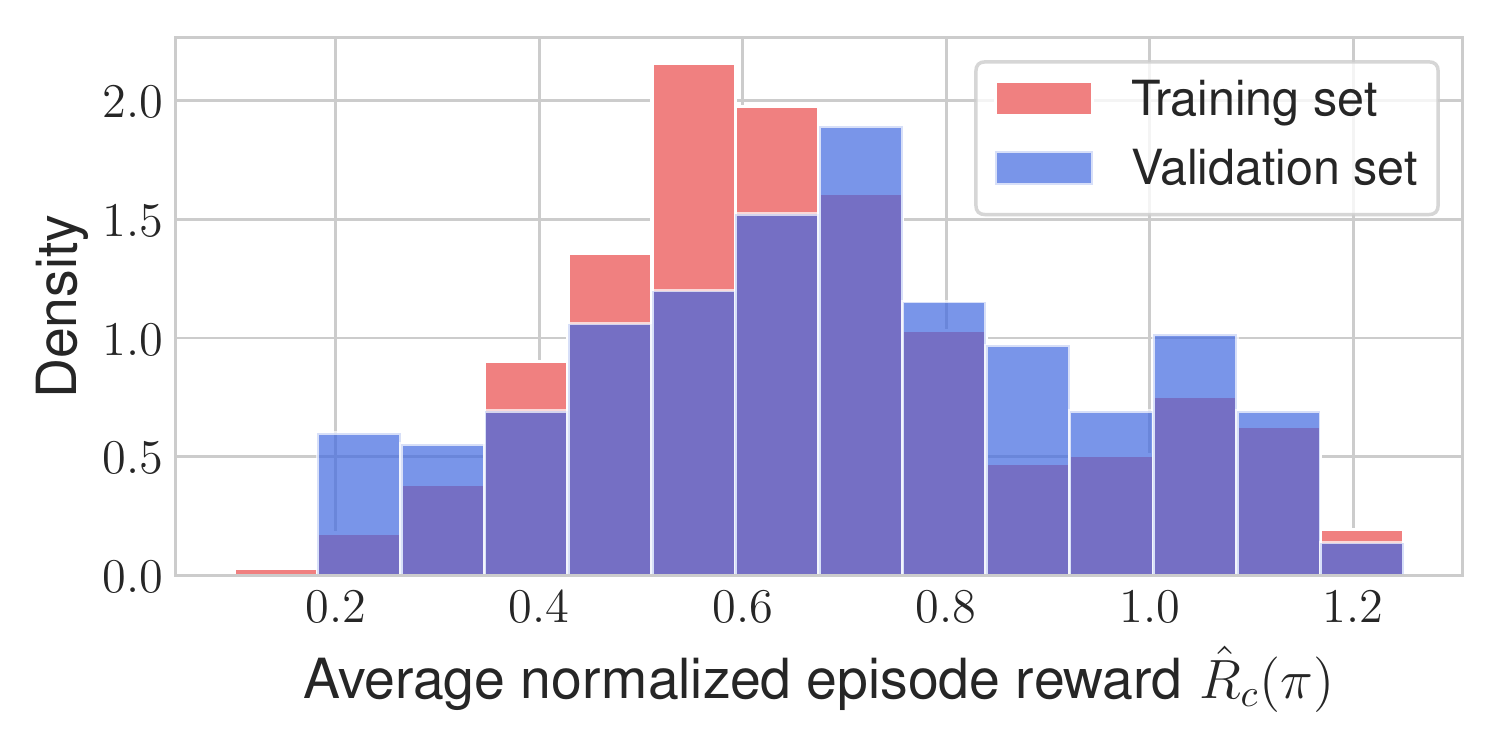}
        \caption{Multi-clip policy's performance on training and validation sets.}
        \label{fig:distillation performance}
    \end{subfigure}
    \hfill
    \begin{subfigure}{0.45\textwidth}
        \includegraphics[width=\linewidth]{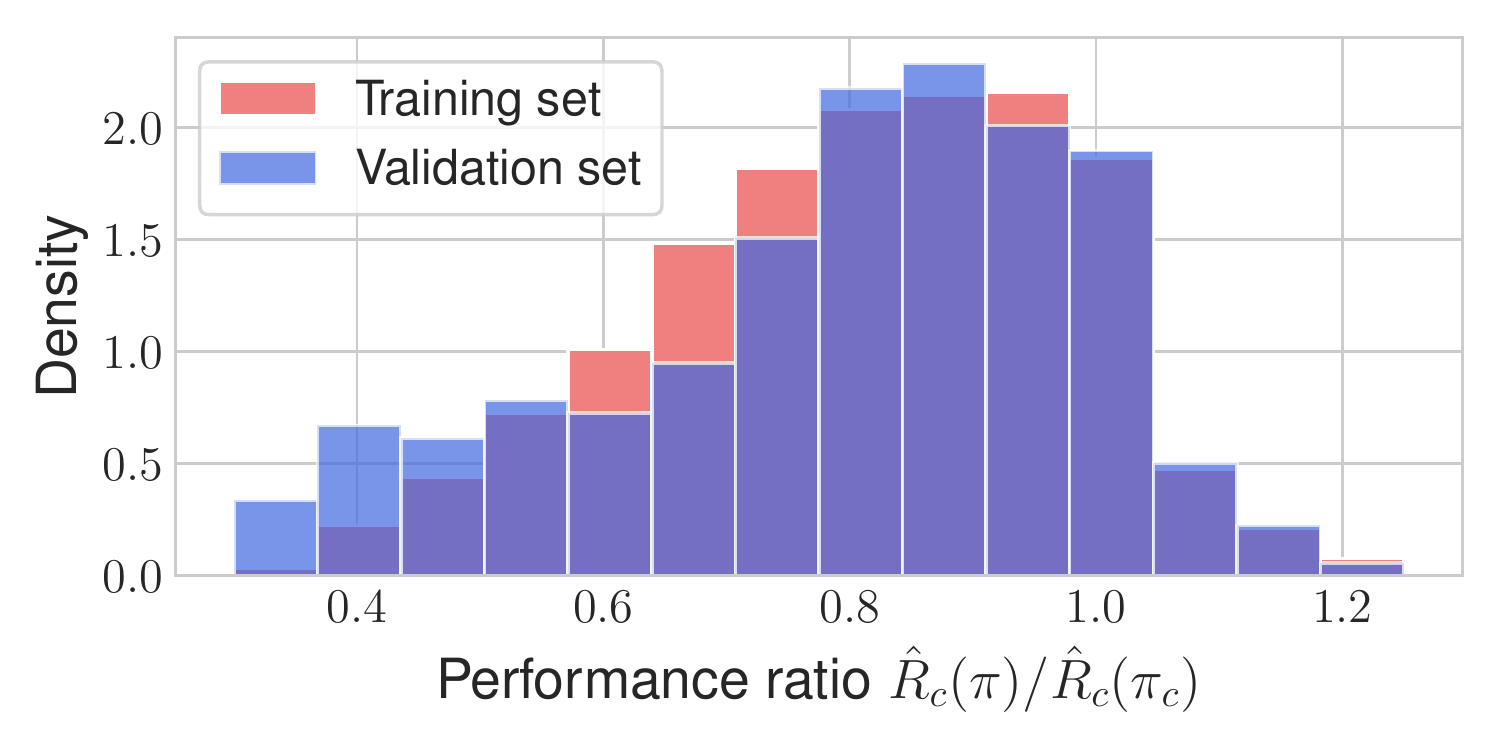}
        \caption{Performance of multi-clip policy relative to expert policies.}
        \label{fig:distillation relative performance}
    \end{subfigure}
    \caption{Performance of RWR-trained multi-clip policy.}
    \label{fig:distillation results}
\end{figure}

To assess the generalization of the multi-clip policy, we train the policy using RWR on a subset of \dataset covering $90\%$ of the \mocap clips.
We treat the remaining $10\%$ of the clips as a validation set when evaluating the multi-clip policy.
We find that the multi-clip policy performs similarly on the training set and validation set clips (\cref{fig:distillation performance}), with the validation set performance even being slightly higher than the training set performance (mean of $0.699$ vs. $0.674$).
This is likely because the clips in the validation set are slightly easier.

To account for the reward scale of the clips, we also report the multi-clip policy's performance relative to the clip experts (\cref{fig:distillation relative performance}).
Again, the training set and validation set relative performances are very similar, though now the multi-clip policy has a small relative performance drop in the validation set (mean of $0.797$ vs. $0.815$).
We also observe that the multi-clip policy outperforms the clip experts on $13\%$ of the \mocap snippets.

We encourage the reader to visit the \projectwebsite to see  videos of the multi-clip policy.

\subsubsection{Re-Use for Reinforcement Learning} \label{sec:re-use for reinforcement learning}
We re-use the decoder $\pi_\decoder$ from an RWR-trained multi-clip policy for reinforcement learning to constrain the behaviors of the humanoid and speed up learning.
In particular, we study two tasks that require adept locomotion skills:
\begin{enumerate}
\item A sparse-reward go-to-target task where the agent receives a non-zero reward only if the humanoid is sufficiently close to the target.
The target relocates once the humanoid stands on it for a few time steps.
\item A velocity control task where shaped rewards encourage the humanoid to go at a given speed in a given direction.
The desired speed and direction change randomly every few seconds.
\end{enumerate}
We treat $\pi_\decoder$ as part of the environment and the motor intention $z$ as the action.
We thus learn a new high-level policy $\pi_\mathrm{task}(z|s)$ that steers the low-level policy to maximize the task reward.

Given the tasks are locomotion-driven, we also consider a more specialized decoder with a 20-dimensional intention which is trained solely on locomotion clips from \dataset (called the ``Locomotion'' subset) to see if further restricting the learned skills offers any more speedup.
As a baseline, we also perform RL without a low-level policy.

\begin{table}[t]
    \footnotesize
    \centering
    \caption{Returns for the transfer tasks, showing the mean and standard deviation over five seeds.}
    \begin{tabular}{r||c|c|c}
    & General low-level policy  & Locomotion low-level policy & No low-level policy \\ \hline
    {\footnotesize Go-to-target} & $96.3 \pm 2.8$ & $66.1 \pm 32.8$ & $7.5 \pm 1.1$ \\ \hline
    {\footnotesize Velocity control} & $1074 \pm 55$ & $884 \pm 81$ & $1157 \pm 89$
    \end{tabular}
    \label{tab:transfer results}
\end{table}

\begin{figure}[t]
    \centering
    \includegraphics[width=0.9\textwidth]{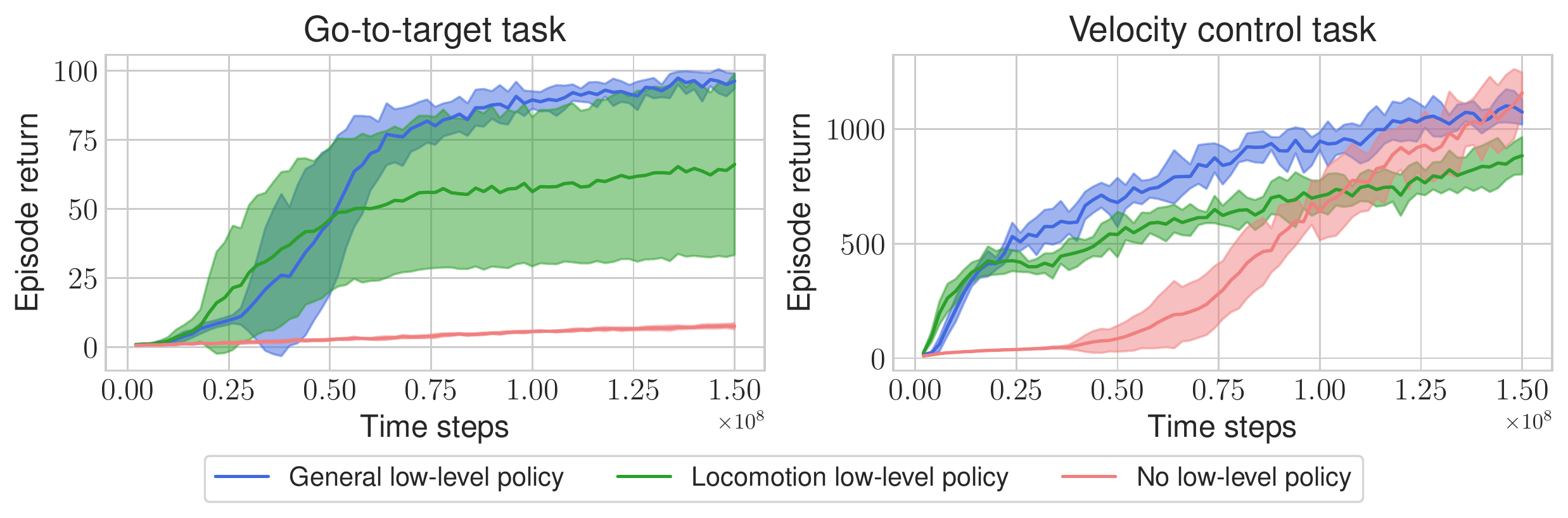}
    \caption{Training curves for transfer tasks.
    All experiments use five seeds.}
    \label{fig:transfer training curves}
\end{figure}

We find that re-using a low-level policy drastically speeds up learning and usually produces higher returns (\cref{tab:transfer results} and \cref{fig:transfer training curves}).
For the go-to-target task, the locomotion-based low-level policy induces faster training than the more general low-level policy, though it does converge to lesser performance and on one out of five seeds converges to a very low reward.
This performance gap is likely a combination of the lower dimensionality of the locomotion policy restricting the degree of control by the high-level policy and the ``Locomotion'' subset excluding some useful behaviors, a result also found by \citet{hasenclever2020comic}.
The baseline without the low-level policy fails to learn the task.
For the velocity control task, the locomotion-based policy induces slightly faster learning than the general policy but again results in lower reward.
The baseline without the low-level policy learns the task more slowly, though it does achieve high reward eventually.

In both tasks, we find that including a pretrained low-level policy produces much more realistic gaits.
The humanoid efficiently runs from target to target in the go-to-target task and smoothly changes speeds and direction of motion in the velocity control task.
On the other hand, the baseline approach produces incredibly unusual motions.
In the go-to-target task, the humanoid convulses and contorts itself towards the first target before falling to the ground.
In the velocity control task, the humanoid rapidly taps the feet to propel the body at the desired velocity.
We encourage the reader to visit the \projectwebsite to see  videos of the RL results.

\subsection{Motion Completion with GPT} \label{sec:gpt}
We also train a GPT model~\citep{radford2019language} based on the minGPT implementation~\citep{kaparthy2020mingpt} to generate motion.
Starting with a motion prompt (sequence of humanoid observations generated by a clip expert), the GPT policy~(\cref{fig:gpt}) autoregressively predicts actions from the context of recent humanoid observations.
We train the GPT by sampling 32-step sequences (corresponding to $1$ second of motion) of humanoid observations $s_{(t-31):t}$ and expert's mean actions $\bar{a}_{(t-31):t}$ from the \dataset dataset $\mathcal D$ and performing supervised learning using the mean squared error loss on the predicted action sequence.

\begin{figure}[t]
    \centering
    \begin{subfigure}{0.4\textwidth}
        \includegraphics[width=\linewidth]{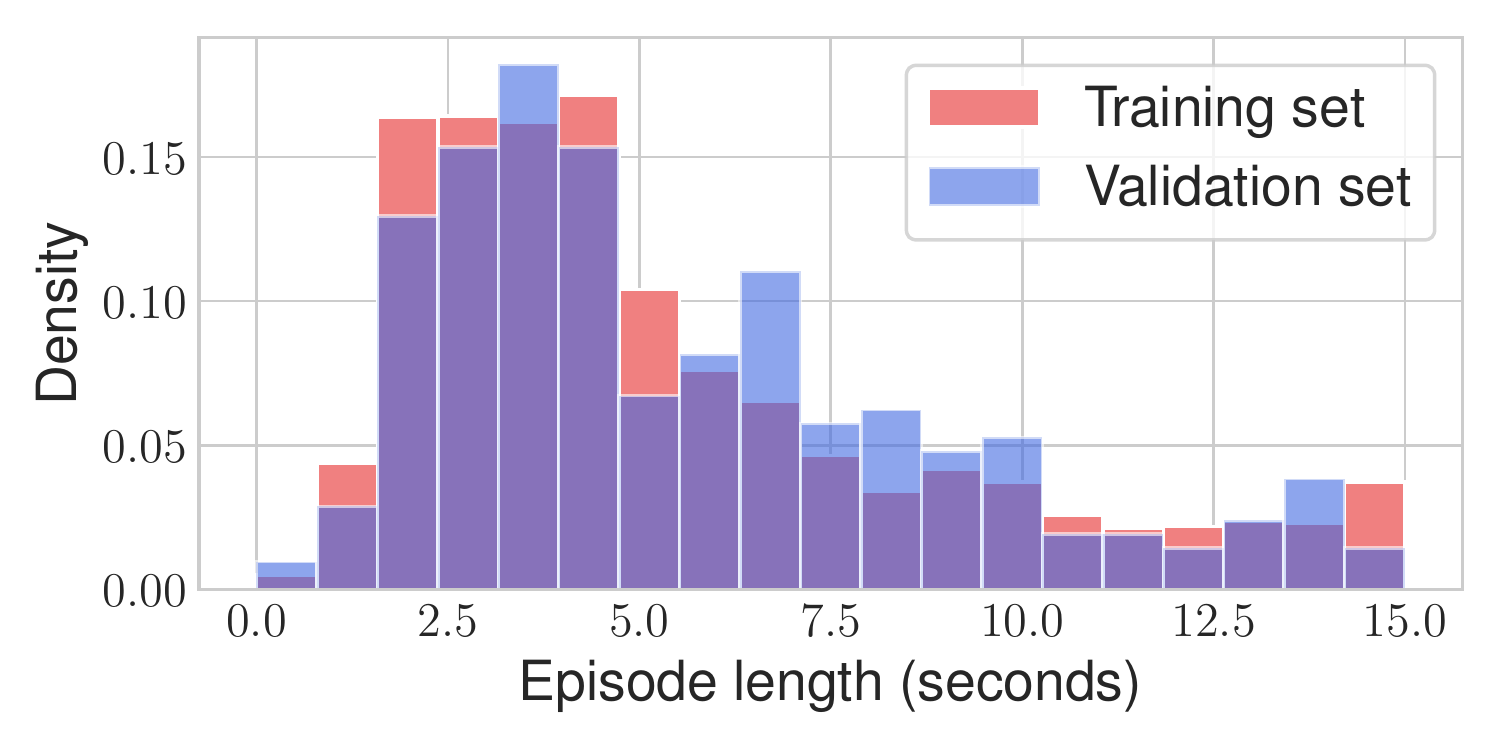}
        \caption{Absolute episode lengths of GPT.}
        \label{fig:gpt episode length}
    \end{subfigure}
    \qquad\qquad
    \begin{subfigure}{0.4\textwidth}
        \includegraphics[width=\linewidth]{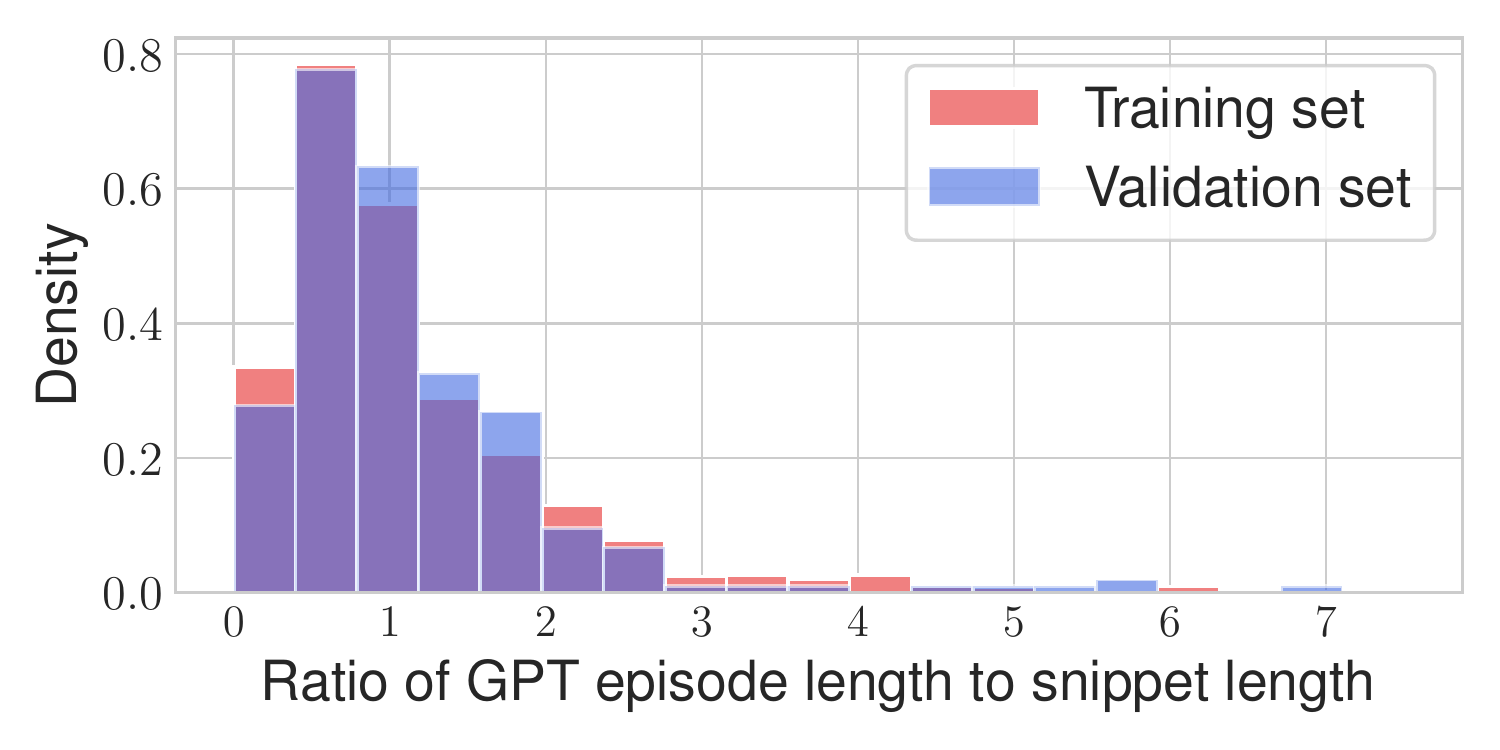}
        \caption{Relative episode lengths of GPT.}
        \label{fig:gpt episode length ratio}
    \end{subfigure}
    \caption{Episode lengths of GPT on \mocap snippets.}
    \label{fig:gpt evaluation}
\end{figure}

To roll out the policy, we provide the GPT policy with a 32-step prompt from a clip expert and let GPT roll out thereafter.
The episode either terminates after 500 steps (about $15$ seconds) or if a body part other than the feet touches the ground (e.g., humanoid falling over).
On many clip snippets, the GPT model is able to control the humanoid for several seconds past the end of the prompt~(\cref{tab:gpt statistics} and \cref{fig:gpt episode length}), with similar lengths on the training set and a held-out validation set of prompts.
We also observe that on many clips the GPT can control the humanoid for several times longer than the length of the corresponding clip snippet~(\cref{tab:gpt statistics} and \cref{fig:gpt episode length ratio}).

\begin{table}[t]
    \footnotesize
    \centering
    \caption{Motion completion statistics on the \mocap snippets.}
    \begin{tabular}{r|c|c|c|c|c}
                              & {\small Mean}    & {\small Standard deviation} & {\small Median} & {\small Minimum} & {\small Maximum} \\ \hline
    Episode length (seconds) & $5.47$ & $3.47$            & $4.38$ & $0.23$ & $15.00$ \\
    Relative episode length  & $1.15$ & $0.94$            & $0.87$ & $0.05$ & $7.63$
    \end{tabular}
    \label{tab:gpt statistics}
\end{table}

\begin{figure}[t!]
    \centering
    \begin{subfigure}{0.25\textwidth}
        \includegraphics[width=\linewidth]{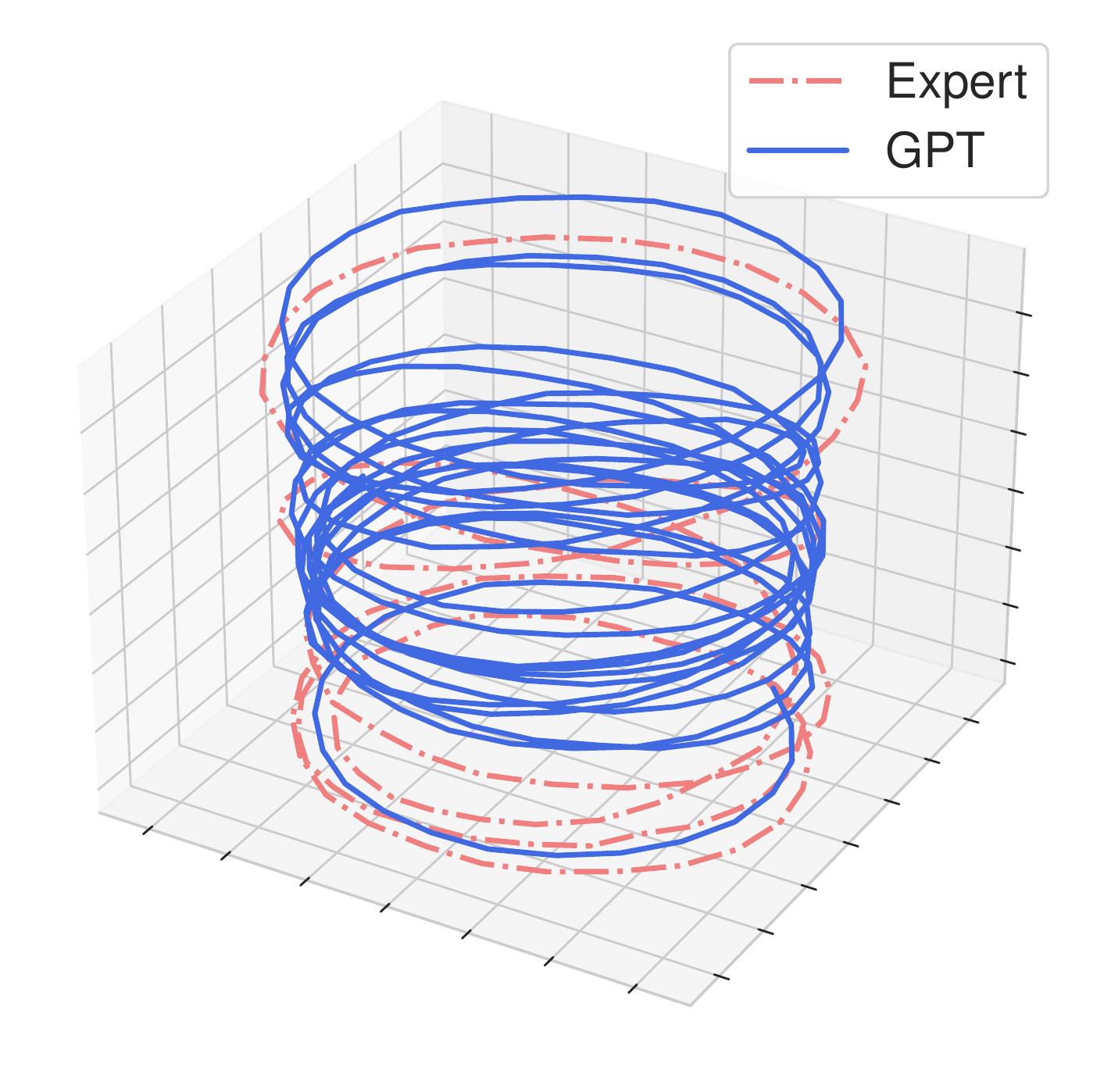}
        \caption{Locomotion clip where behaviors align.}
        \label{fig:gpt same locomotion}
    \end{subfigure}
    \hfill
    \begin{subfigure}{0.25\textwidth}
        \includegraphics[width=\linewidth]{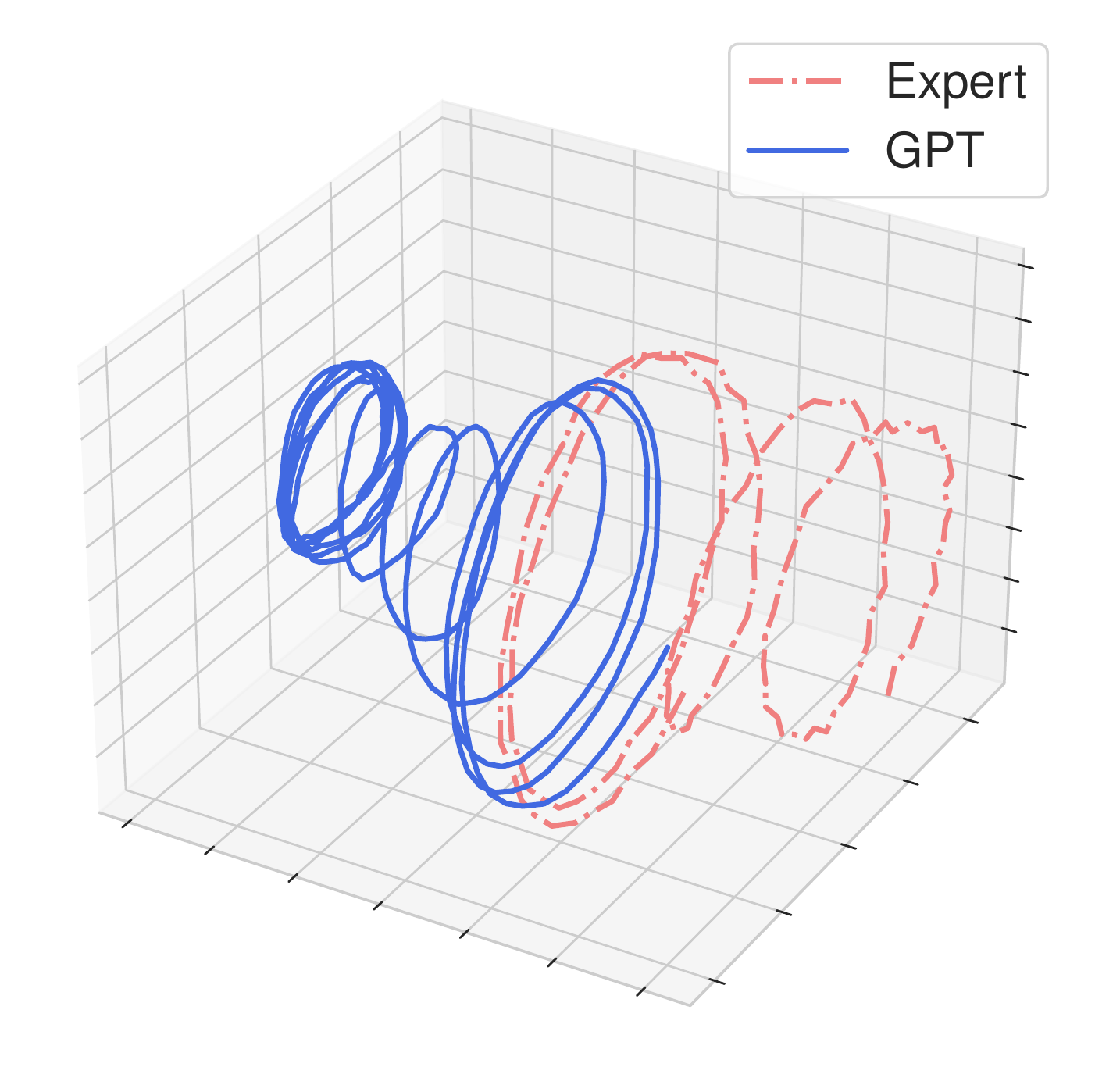}
        \caption{Locomotion clip where behaviors differ.}
        \label{fig:gpt different locomotion}
    \end{subfigure}
    \hfill
    \begin{subfigure}{0.25\textwidth}
        \includegraphics[width=\linewidth]{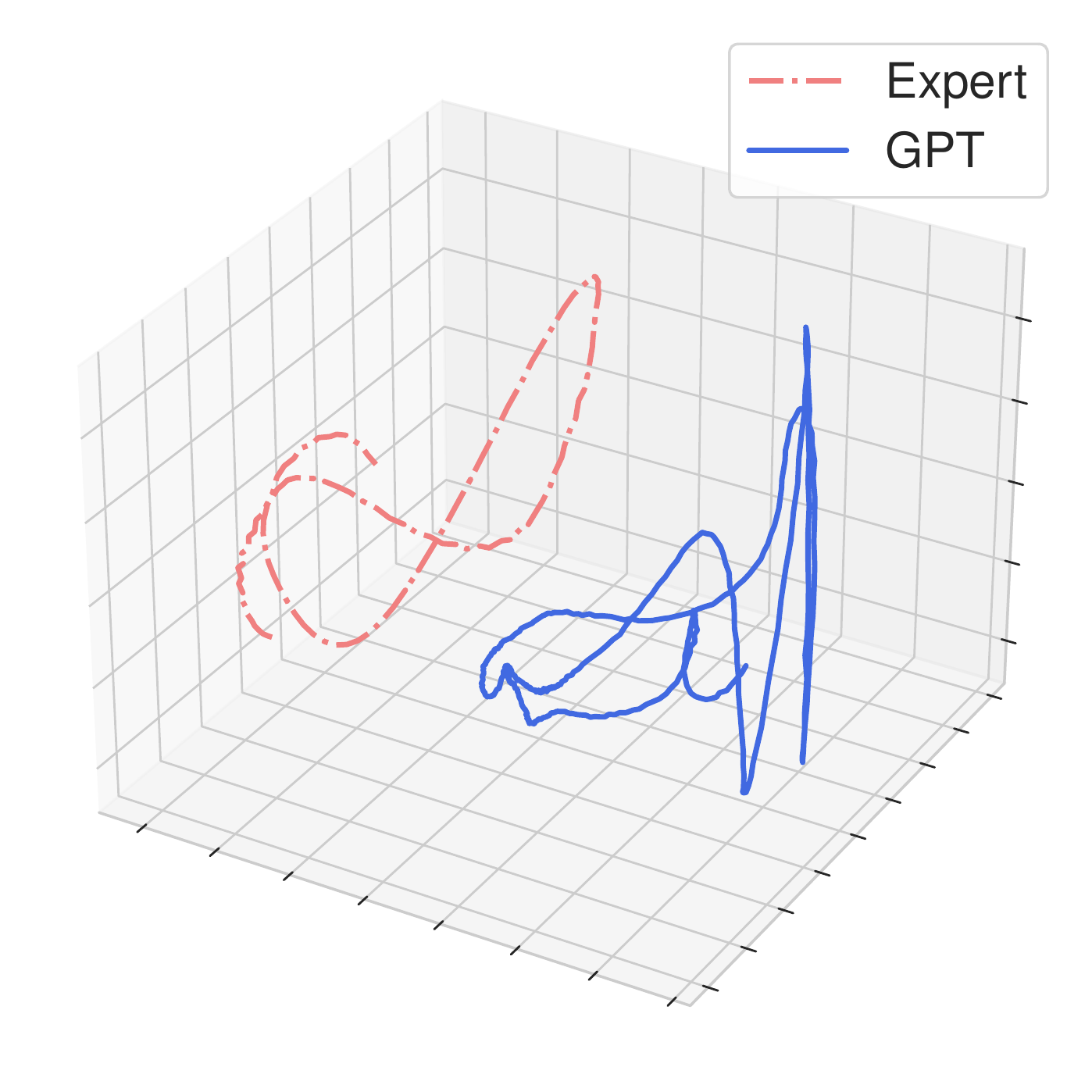}
        \caption{Non-locomotion clip where behaviors differ.}
        \label{fig:gpt different behavior}
    \end{subfigure}
    \caption{PCA projections of action sequences of length $32$ from experts and GPT.}
    \label{fig:pca projections}
\end{figure}

To visualize the rollouts, we perform principal component analysis~(PCA) on action sequences of length $32$ applied by GPT and the snippet expert used to generate the motion prompt~(\cref{fig:pca projections}).
Qualitatively, we find that GPT usually repeats motions demonstrated in locomotion prompts, such as the running motion corresponding to~\cref{fig:gpt same locomotion}.
Occasionally, GPT will produce a different motion than the underlying clip, usually due to ambiguity in the prompt.
For example, in~\cref{fig:gpt different locomotion}, GPT has the humanoid repeatedly step backwards, whereas the expert takes repeated side steps.
In~\cref{fig:gpt different behavior}, the GPT policy performs an entirely different arm-waving motion than that of the expert.
We encourage the reader to visit the \projectwebsite to see videos of GPT motion completion.

\section{Discussion}\label{sec:disc}

We presented a dataset of high-quality \mocap-tracking policies and their rollouts for the \dmcontrol humanoid environment.
From these rollouts, we trained multi-clip tracking
policies that can be re-used for new high-level tasks and GPT policies which can generate humanoid motion when given a prompt.
We have open sourced our dataset, models, and code under permissive licenses.

We do point out that our models and data are only applicable to the \dmcontrol environment, which uses MuJoCo as the backend simulator.
We also point out that all considered clips only occur on flat ground and do not include any human or object interaction.
Though this seems to limit the environments and tasks where this dataset is applicable, the \dmcontrol package~\citep{tunyasuvunakool2020dm_control} has tools to change the terrain, add more \mocap clips, and add objects (e.g., balls) to the environment.
Indeed, prior work has used custom clips which include extra objects~\citep{merel2020catch,liu2022motor}.
While the dataset and domain may raise concerns on automation, we believe the considered simulated domain is limited enough to not be of ethical import.

This work significantly lowers the barrier of entry for simulated humanoid control, which promises to be a rich field for studying multi-task learning and motor intelligence.
In addition to the showcases presented, we believe this dataset can be used in training other policy architectures like decision and trajectory transformers~\citep{chen2021decision,janner2021offline} or in setups like offline reinforcement learning~\citep{fu2020d4rl,levine2020offline} as the dataset allows research groups to bypass the time- and energy-consuming process of learning low-level motor skills from \mocap data.

\begin{ack}
We thank Leonard Hasenclever for providing helpful information used in DeepMind's prior work on humanoid control.
We also thank Byron Boots for suggesting to use PCA projections for visualization.
Finally, we thank the reviewers for their invaluable feedback.

The data used in this project was obtained from \mbox{\url{mocap.cs.cmu.edu}}.
The database was created with funding from NSF EIA-0196217.
\end{ack}

\bibliography{references}
\bibliographystyle{abbrvnat}

\newpage

\begin{appendix}

This appendix provides the following sections:
\begin{itemize}
    \item documentation of the dataset (\cref{app:dataset}),
    \item training details (\cref{app:training details}), and
    \item extra results (\cref{app:results}).
\end{itemize}

\section{Dataset Documentation} \label{app:dataset}

\subsection{Clip Snippet Experts} \label{app:clip expert descriptions}
We signify a clip snippet expert by the snippet it is tracking.
We denote a snippet by the clip ID, its start step, and its end step.
For example, \texttt{CMU\char`_006\char`_12-151-336} is the snippet corresponding to the clip \texttt{CMU\char`_006\char`_12} with start step $151$ and end step $336$.
Taking \texttt{CMU\char`_006\char`_12-151-336} as an example expert, the file hierarchy for the snippet expert is: \\
\dirtree{%
.1 CMU\char`_006\char`_12-151-336.
.2 clip\char`_info.json\DTcomment{Contains clip ID, start step, and end step.}.
.2 eval\char`_rsi/model.
    .3 best\char`_model.zip\DTcomment{Contains policy parameters and hyperparameters.}.
    .3 vecnormalize.pkl\DTcomment{Used to get normalizer for observation and reward.}.
}
The expert policy can be loaded using Stable-Baselines3's functionality.

\subsection{Expert Rollouts} \label{app:expert rollout descriptions}
The expert rollouts consist of a collection of HDF5 files, one file per clip.
An HDF5 file contains expert rollouts for each constituent snippet as well as miscellaneous information and statistics.
To facilitate efficient loading of the observations, we concatenate all the proprioceptive observations (joint angles, joint velocities, actuator activations, etc.) from an episode into a single numerical array and provide indices for the constituent observations in the \texttt{observable\char`_indices} group.

Taking \texttt{CMU\char`_009\char`_12.hdf5} (which contains three snippets) as an example, we have the following HDF5 hierarchy: \\
\dirtree{%
.1 CMU\char`_009\char`_12.hdf5.
.2 n\char`_rsi\char`_rollouts\DTcomment{$R$, number of rollouts from random time steps in snippet.}.
.2 n\char`_start\char`_rollouts\DTcomment{$S$, number of rollouts from start of snippet.}.
.2 ref\char`_steps\DTcomment{\small Indices of \mocap reference relative to current time step. Here, $(1,2,3,4,5)$.}.
.2 observable\char`_indices.
    .3 walker.
        .4 actuator\char`_activation\DTcomment{$(0, 1, \ldots, 54, 55)$}.
        .4 appendages\char`_pos\DTcomment{$(56, 57, \ldots, 69, 70)$}.
        .4 body\char`_height\DTcomment{$(71)$}.
        .4 $\vdots$.
        .4 world\char`_zaxis\DTcomment{$(2865, 2866, 2867)$}\\.
.2 stats\DTcomment{Statistics computed over the entire dataset.}.
    .3 act\char`_mean\DTcomment{Mean of the experts' sampled actions.}.
    .3 act\char`_var\DTcomment{Variance of the experts' sampled actions.}.
    .3 mean\char`_act\char`_mean\DTcomment{Mean of the experts' mean actions.}.
    .3 mean\char`_act\char`_var\DTcomment{Variance of the experts' mean actions.}.
    .3 proprio\char`_mean\DTcomment{Mean of the proprioceptive observations.}.
    .3 proprio\char`_var\DTcomment{Variance of the proprioceptive observations.}.
    .3 count \DTcomment{Number of observations in dataset.}\\.
.2 CMU\char`_009\char`_12-0-198\DTcomment{Rollouts for the snippet \texttt{CMU\char`_009\char`_12-0-198}.}.
.2 CMU\char`_009\char`_12-165-363\DTcomment{Rollouts for the snippet \texttt{CMU\char`_009\char`_12-165-363}.}.
.2 CMU\char`_009\char`_12-330-529\DTcomment{Rollouts for the snippet \texttt{CMU\char`_009\char`_12-330-529}.}.
}

Each snippet group contains $R+S$ rollouts.
The first $S$ episodes correspond to episodes intialized from the start of the snippet and the last $R$ episodes to episodes initialized at random points in the snippet.
We now uncollapse the \texttt{CMU\char`_009\char`_12-0-198} group within the HDF5 file to reveal the rollout structure: \\
\dirtree{%
.1 CMU\char`_009\char`_12-0-198.
    .2 early\char`_termination\DTcomment{\small $(R+S)$-boolean array indicating which episodes terminated early.}\\.
    .2 rsi\char`_metrics\DTcomment{Metrics for episodes that initialize at random points in snippet.}.
        .3 episode\char`_returns\DTcomment{$R$-array of episode returns.}.
        .3 episode\char`_lengths\DTcomment{$R$-array of episode lengths.}.
        .3 norm\char`_episode\char`_returns\DTcomment{$R$-array of normalized episode rewards.}.
        .3 norm\char`_episode\char`_lengths\DTcomment{$R$-array of normalized episode lengths.}\\.
    .2 start\char`_metrics\DTcomment{Metrics for episodes that initialize at start in snippet.}\\.
    .2 0\DTcomment{First episode, of length $T$.}.
        .3 observations.
            .4 proprioceptive\DTcomment{$(T+1)$-array of proprioceptive observations.}.
            .4 walker/body\char`_camera\DTcomment{\small $(T+1)$-array of images from body camera (not included).}.
            .4 walker/egocentric\char`_camera\DTcomment{\scriptsize $(T+1)$-array of images from egocentric camera (not included).}\\.
        .3 actions\DTcomment{$T$-array of sampled actions executed in environment.}.
        .3 mean\char`_actions\DTcomment{$T$-array of corresponding mean actions.}.
        .3 rewards\DTcomment{$T$-array of rewards from environment.}.
        .3 values\DTcomment{$T$-array computed using the policy's value network.}.
        .3 advantages\DTcomment{$T$-array computed using generalized advantage estimation.}\\.
    .2 1\DTcomment{Second episode.}.
    .2 $\vdots$.
    .2 $R+S-1$\DTcomment{$(R+S)^\mathrm{th}$ episode.}.
}
To keep the dataset size manageable, we do \emph{not} include image observations in the dataset.
We do provide code to log them when rolling out the experts for generating the dataset.

\subsection{Hosting Plan} \label{sec:hosting plan}
The link to the dataset can be found on the \projectwebsite.
We provide a ``large'' rollout dataset where $R = S = 100$ with size $600~\mathrm{GB}$ and a ``small'' rollout dataset where $R = S = 10$ with size $50~\mathrm{GB}$.
The dataset website also includes the policies we trained in~\cref{sec:applications}, i.e., the multi-clip tracking policies, RL-trained task policies, and the GPT policy.
We also provide a Python script to download individual experts and rollouts from the dataset.

\newpage
\section{Training Details} \label{app:training details}
\subsection{Clip Snippet Experts} \label{app:clip experts}
\subsubsection{\mocap Snippets} \label{app:mocap snippets}
\begin{figure}[h!]
    \centering
    \begin{subfigure}{0.45\textwidth}
        \includegraphics[width=\linewidth]{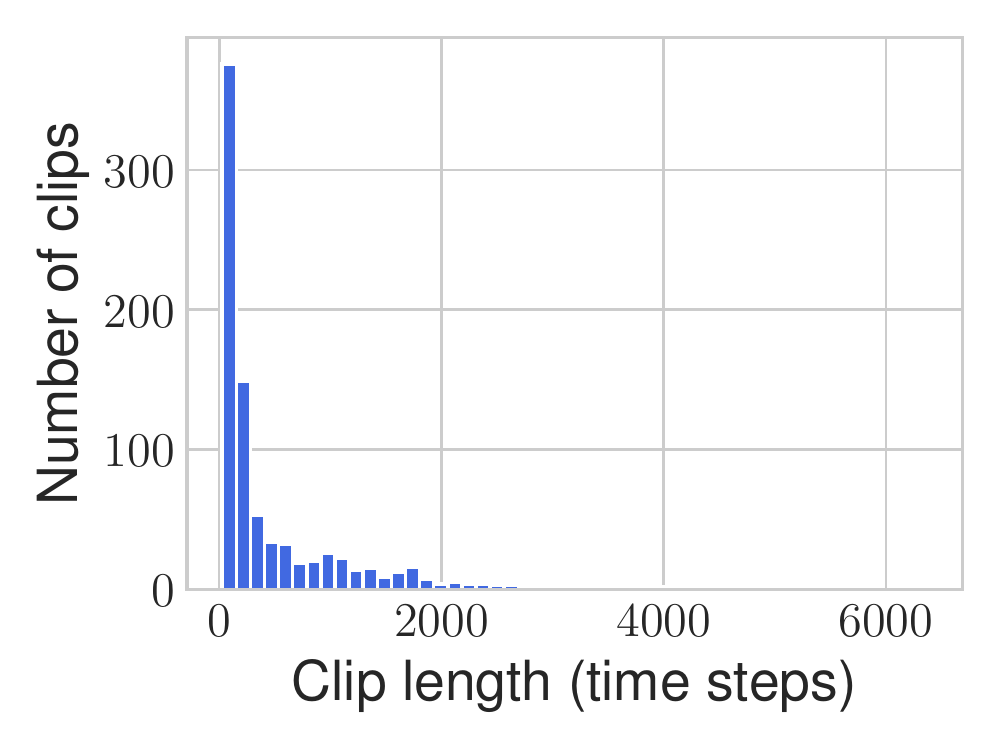}
        \caption{Lengths of the \mocap clips.}
        \label{fig:clip lengths}
    \end{subfigure}
    \hfill
    \begin{subfigure}{0.45\textwidth}
        \includegraphics[width=\linewidth]{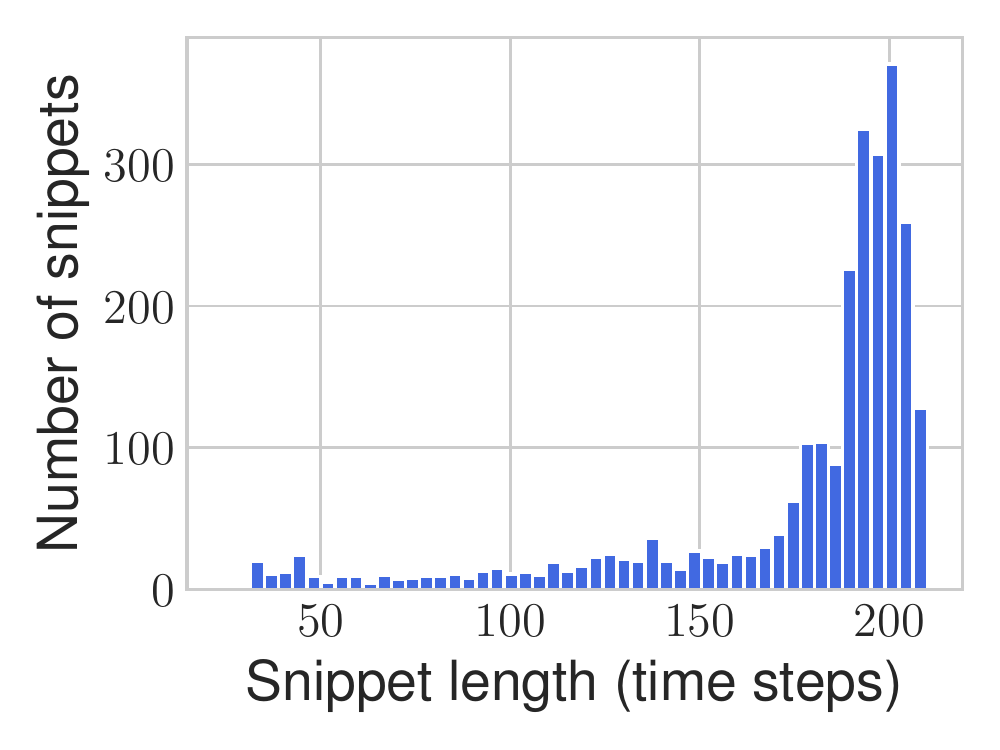}
        \caption{Lengths of the snippets generated from the clips.}
        \label{fig:snippet lengths}
    \end{subfigure}
    \caption{Lengths of clips and snippets.}
    \label{fig:clip division}
\end{figure}

The \mocap dataset has a wide spread in clip length (\cref{fig:clip lengths}), with the longest clip being $6371$ time steps ($191$ seconds).
Training clip experts to track long clips is potentially slow and laborious, so we follow \citet{merel2019neural} by dividing clips longer than $210$ time steps ($6.3$ seconds) into short snippets.
In particular, we divide the clip into uniformly-sized snippets with an overlap of $33$ time steps ($1$ second) such that the longest snippet has at most $210$ time steps.
This yields a snippet dataset with a much tighter range of snippet lengths (\cref{fig:snippet lengths}).
We do \emph{not} divide the clips from the ``Get Up'' subset of the \mocap dataset since they contain involved motions of getting up from the ground.

\subsubsection{Expert Training Details} \label{app:clip expert training}
\begin{table}[h!]
    \centering
    \caption{Hyperparameters for clip snippet expert training.}
    \begin{tabular}{|c|c|} \hline
    Total environment steps & \num{150} million \\ \hline
    Environment steps per policy update & \num{8192} \\ \hline
    PPO epochs & \num{10} \\ \hline
    PPO minibatch size & \num{512} \\ \hline
    PPO clipping parameter $\varepsilon$ & \num{0.25} \\ \hline
    GAE parameter $\lambda$ & \num{0.95} \\ \hline
    Discount factor $\gamma$ & \num{0.95} \\ \hline
    $\ell_2$ gradient norm clipping value & \num{1} \\ \hline
    Adam step size & \shortstack{\num{1e-5} for first $50\mathrm{M}$ env. steps \\ \num{6e-6} for next $50\mathrm{M}$ env. steps \\ \num{3e-6} for last $50\mathrm{M}$ env. steps} \\ \hline
    \end{tabular}
    \label{tab:clip expert hyperparameters}
\end{table}

We use the Stable-Baselines3~\citep{raffin2021stable} implementation of PPO~\citep{schulman2017proximal} to optimize each expert.
Each expert is a neural network with three hidden layers, 1024 neurons in each hidden layer, and the $\tanh$ activation.
At the start of each episode, we randomly select a time step from the corresponding clip snippet (excluding the last 10 time steps from the snippet) and initialize the humanoid to match the clip features at the corresponding step.
We evaluate the policy every 1 million environment steps using 1000 episodes under the same initialization scheme (but now excluding the last 30 time steps of the snippet) and the same action noise of $0.1$ as for training rollouts.
We also end the training if the average normalized episode length is at least $0.98$ and the average normalized episode reward does not improve by more than $1\%$ from the current best reward after 10 million environment steps.
We normalize the observation and reward using running statistics from the environment.
We give the other relevant hyperparameters in \cref{tab:clip expert hyperparameters}.
For details of the reward function and early termination of an episode, we refer the reader to the appendix of \citet{hasenclever2020comic}.

We ran the training on a mix of Azure Standard\_H8 (8 CPUs), Standard\_H16 (16 CPUs), Standard\_NC6s\_v2 (6 CPUs and 1 P100 GPU), and Standard\_ND6s (6 CPUs and 1 P40 GPU) VMs.

The observables for the clip expert are:
\texttt{joints\char`_pos}, \texttt{joints\char`_vel}, \texttt{sensors\char`_velocimeter}, \texttt{sensors\char`_gyro}, \texttt{end\char`_effectors\char`_pos}, \texttt{world\char`_zaxis}, \texttt{actuator\char`_activation}, \texttt{sensors\char`_touch}, \texttt{sensors\char`_torque}, \texttt{time\char`_in\char`_clip}.

\subsection{Multi-Clip Tracking Policy} \label{app:distillation}
\begin{table}[h!]
    \centering
    \caption{Hyperparameters for multi-clip tracking policy training.}
    \begin{tabular}{|c|c|} \hline
    Adam step size & \num{5e-4} \\ \hline
    Minibatch sequence length $T$ & \num{30} \\ \hline
    Minibatch size & \num{256} \\ \hline
    $\ell_2$ gradient norm clipping value & \num{1} \\ \hline
    KL divergence weight $\beta$ & \num{0.1} \\ \hline
    Autoregressive parameter $\alpha$ & \num{0} \\ \hline
    Weighting temperature $\lambda$ & \shortstack{CWR: \num{0.2} \\ AWR: \num{8} \\ RWR: \num{4} } \\ \hline
    \end{tabular}
    \label{tab:multi-clip policy hyperparameters}
\end{table}

We train the multi-clip policy $\pi(a_t, z_t | s_t, s_t^\reference, z_{t-1}) = \pi_\encoder(z_t | s_t, s_t^\reference, z_{t-1}) \pi_\decoder(a_t | s_t, z_t)$ by optimizing the following imitation objective:
\begin{small}
\[
\mathbb{E}_{\subalign{(s_{1:T}, s_{1:T}^\reference, \bar{a}_{1:T}, c) &\sim \mathcal{D}, \\ z_{0:T} &\sim \pi_\encoder}} \left[ \sum_{t=1}^T \left[ w_c(s_t, \bar{a}_t) \log \pi_\decoder(\bar{a}_t | s_t, z_t) - \beta \, \mathrm{KL}(\pi_\encoder (z_t | s_t, s_t^\reference, z_{t-1}) \,\|\, p(z_t | z_{t-1})) \right] \right],
\]
\end{small}%
where $p(z_t|z_{t-1}) = \mathcal{N}(z_t; \alpha z_{t-1}, (1-\alpha^2) I)$ for some $\alpha \in [0, 1]$.
We do this (for each data point in a minibatch) by sampling $z_0 \sim \mathcal{N}(0, I)$, sampling a $T$-step data sequence (of humanoid states $s_{1:T}$, \mocap references $s_{1:T}^\reference$, and expert's mean actions $\bar{a}_{1:T}$) from the dataset $\mathcal{D}$, unrolling the recurrent policy through the sampled sequence, performing backpropagation through time on the objective function, and finally updating the network using the Adam optimizer~\citep{kingma2015adam}.
To speed up training, we normalize the humanoid state $s_t$ and \mocap reference $s_t^\reference$ using the corresponding mean and standard deviation computed over the entire dataset.
For the weighted schemes, we multiply the weight $w_c$ by a constant that ensures the average data weight is $1$ so that the KL regularization term maintains the same relative weight.
For all schemes, we also sample data from shorter clips at a higher rate to ensure the rollout data from the clips is uniformly even.
This gives about $1\%$ improvement in policy evaluation compared to vanilla sampling.

We use PyTorch Lightning~\citep{falcon2019pytorch} to train the multi-clip policy.
The encoder and decoder are both neural networks with $1024$ neurons per hidden layer and use layer norm and the $\mathrm{ELU}$ activation.
The encoder has two hidden layers, while the decoder has three hidden layers.
We ran the training on Azure Standard\_ND24s VMs, each equipped with 24 CPUs and 4 P40 GPUs.
We periodically evaluate the multi-clip policy by running 1000 episodes on the set of \mocap snippets following the same reference state initialization scheme as in the rest of the paper.
We found we only need to train the policy for about \num{50000} steps (about $10\%$ of an epoch) before plateauing on policy evaluation~(\cref{sec:multi-clip training curves}).
We give the other relevant hyperparameters in \cref{tab:multi-clip policy hyperparameters}.

The observables for the policy are:
\begin{itemize}
\item Encoder: \texttt{joints\char`_pos}, \texttt{joints\char`_vel}, \texttt{sensors\char`_velocimeter}, \texttt{sensors\char`_gyro}, \texttt{end\char`_effectors\char`_pos}, \texttt{world\char`_zaxis}, \texttt{actuator\char`_activation}, \texttt{sensors\char`_touch}, \texttt{sensors\char`_torque}, \texttt{body\char`_height}, \texttt{reference\char`_rel\char`_bodies\char`_pos\char`_local}, \texttt{reference\char`_rel\char`_bodies\char`_quats}
\item Decoder: \texttt{joints\char`_pos}, \texttt{joints\char`_vel}, \texttt{sensors\char`_velocimeter}, \texttt{sensors\char`_gyro}, \texttt{end\char`_effectors\char`_pos}, \texttt{world\char`_zaxis}, \texttt{actuator\char`_activation}, \texttt{sensors\char`_touch}, \texttt{sensors\char`_torque}
\end{itemize}

\subsection{Transfer for Reinforcement Learning} \label{app:transfer}

\subsubsection{Go-to-Target Task}
This task matches that of~\citet{hasenclever2020comic}, which we refer the reader to for details.

\subsubsection{Velocity Control}
In this task, a target speed $s^* \in [0, 4.5]$ and direction $\psi^* \in [0, 2\pi)$ are randomly sampled every $10$ seconds.
Defining the target velocity as $v_t^* = (s^* \cos\psi^*, s^* \sin\psi^*)$ and the humanoid's current velocity as $v_t$, the reward is defined as the product of a speed factor and direction factor:
\[
r_t = \exp\left(-\left(\frac{\|v_t\| - \|v_t^*\|}{\eta}\right)^2 \right) \left(\frac{1 + \mathrm{score}(v_t, v_t^*)}{2} \right)^k,
\]
where $\mathrm{score}(v_t, v_t^*) = v_t \cdot v_t^* / \|v_t\| \|v_t^*\|$ gives the cosine of the angle between the two velocity vectors.
In our experiments, we set $\eta = 0.75$ and $k = 7$.
We also experimented with the velocity error reward used by~\citet{bohez2022imitate} but found that our reward was easier to optimize.
We terminate the episode either after 2000 time steps ($60$ seconds) or if any body part other than the feet touches the ground.

\subsubsection{Hyperparameters}

\begin{table}[h!]
    \centering
    \caption{Hyperparameters for RL transfer tasks.}
    \begin{tabular}{|c|c|} \hline
    Total environment steps & \num{150} million \\ \hline
    Environment steps per policy update & \num{16384} \\ \hline
    PPO epochs & \num{10} \\ \hline
    PPO minibatch size & \num{1024} \\ \hline
    PPO clipping parameter $\varepsilon$ & \num{0.2} \\ \hline
    KL divergence threshold for early stopping & \num{0.3} \\ \hline
    Entropy bonus coefficient & \shortstack{General low-level policy: \num{1e-4} \\ Locomotion low-level policy: \num{1e-3} \\ No low-level policy: \num{1e-4}} \\ \hline
    GAE parameter $\lambda$ & \num{0.95} \\ \hline
    Discount factor $\gamma$ & \num{0.99} \\ \hline
    $\ell_2$ gradient norm clipping value & \num{1} \\ \hline
    Adam step size & \num{5e-5} \\ \hline
    Number of actors & \num{32} \\ \hline
    Initial standard deviation for task policy & \shortstack{With low-level policy: \num{2.5} \\ Without low-level policy: \num{0.5}} \\ \hline
    Maximum per-element action magnitude for task policy & \shortstack{With low-level policy: \num{3} \\ Without low-level policy: \num{1}} \\ \hline
    \end{tabular}
    \label{tab:RL transfer hyperparameters}
\end{table}

Like the snippet experts, we train the task policies using the Stable-Baselines3 implementation of PPO.
Each task policy is a neural network with three hidden layers, 1024 neurons per hidden layer, and the $\tanh$ activation.
We ran the training on Azure Standard\_ND6s (6 CPUs and 1 NVIDIA P40 GPU) VMs.
We give other hyperparameters in~\cref{tab:RL transfer hyperparameters}.

\subsection{Motion Completion with GPT} \label{app:gpt}

\begin{table}[h!]
    \centering
    \caption{Hyperparameters for GPT training.}
    \begin{tabular}{|c|c|} \hline
    Adam step size & \num{3e-6} \\ \hline
    Minibatch size & \num{256} \\ \hline
    $\ell_2$ gradient norm clipping value & \num{1} \\ \hline
    Attention dropout probability & \num{0.1} \\ \hline
    Embedding dropout probability & \num{0.1} \\ \hline
    Residual dropout probability & \num{0.1} \\ \hline
    Block size & \num{32} \\ \hline
    Embedding size & \num{768} \\ \hline
    Attention heads & \num{8} \\ \hline
    Number of layers & \num{8} \\ \hline
    Weight decay & \num{0.1} \\ \hline
    \end{tabular}
    \label{tab:GPT hyperparameters}
\end{table}

We train a variant of minGPT~\citep{kaparthy2020mingpt} that we adapted to accept continuous inputs and output continuous actions.
This particular model has 57 million parameters and was trained with a context length of 32 time steps, corresponding to roughly one second of motion.
Similar to the multi-clip policy~(\cref{app:distillation}), we sample state $s_{(t-31):t}$ and mean-action $\bar{a}_{(t-31):t}$ sequences of length $32$ from the \dataset dataset $\mathcal{D}$.
To speed up training, we normalize the humanoid state $s_t$ using the corresponding mean and standard deviation computed over the entire dataset.
We use the mean squared error loss on the sequence of predicted actions from the GPT.
We trained GPT using PyTorch Lightning~\citep{falcon2019pytorch} on Azure Standard\_NC24s\_v3, each equipped with 24 CPUs and 4 V100 GPUs, for 2 million steps, corresponding to one week of wall-clock time.
We give the other relevant hyperparameters in \cref{tab:GPT hyperparameters}.

The observables for the GPT policy are: \texttt{joints\char`_pos}, \texttt{joints\char`_vel}, \texttt{sensors\char`_velocimeter}, \texttt{sensors\char`_gyro}, \texttt{end\char`_effectors\char`_pos}, \texttt{world\char`_zaxis}, \texttt{actuator\char`_activation}, \texttt{sensors\char`_touch}, \texttt{sensors\char`_torque}, \texttt{body\char`_height}.
Importantly, GPT is not given any reference data from the \mocap clip, so any motion generation was done only on the basis of the historical context provided.

\newpage

\section{More Results} \label{app:results}
\subsection{Clip Snippet Experts}
\subsubsection{Training Curves}
\begin{figure}[ht!]
    \centering
    \includegraphics[width=0.7\textwidth]{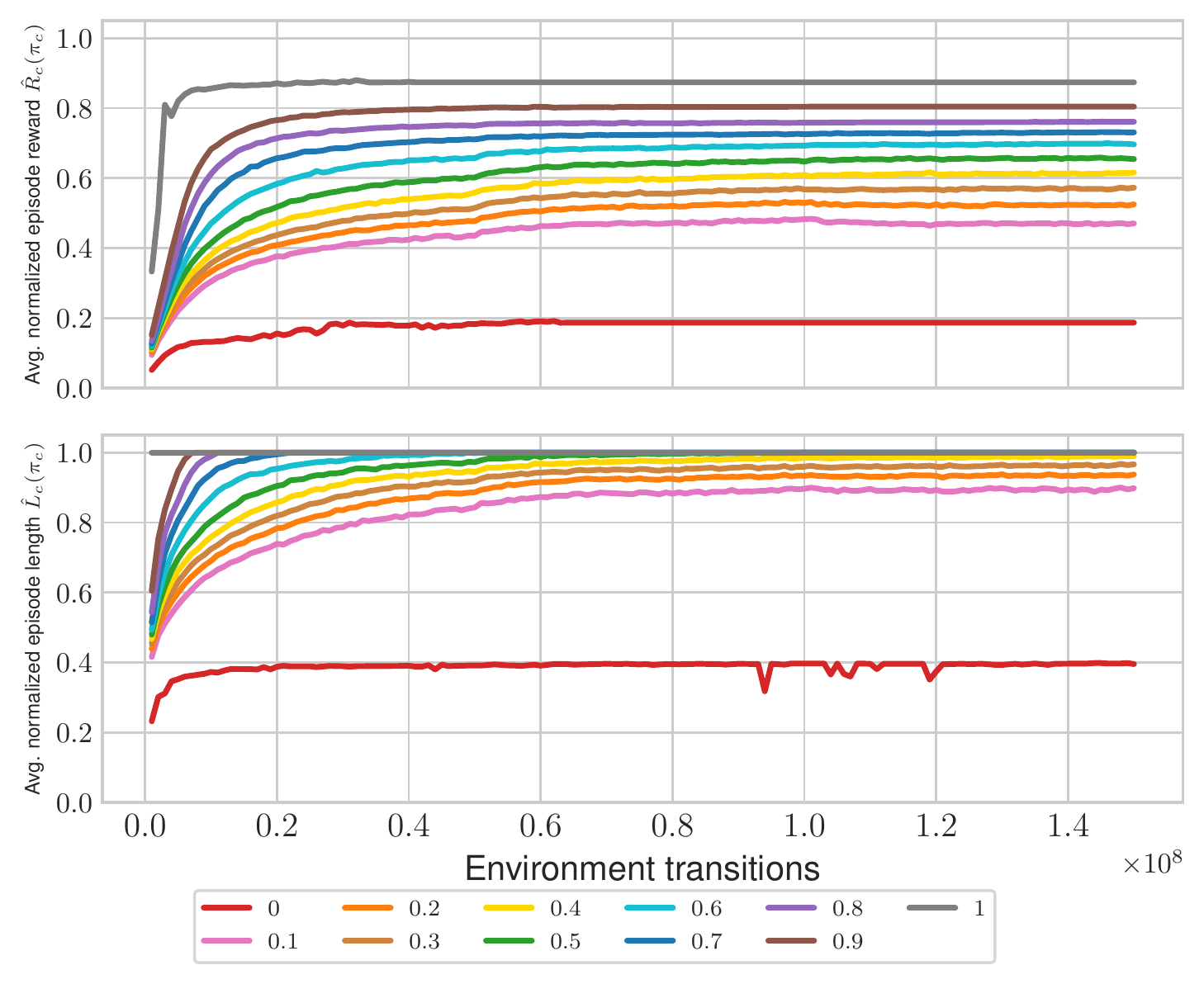}
    \caption{Snippet expert training curves on \mocap dataset.}
    \label{fig:expert learning curves}
\end{figure}
We give the learning curves for the experts in~\cref{fig:expert learning curves}.
In particular, we plot the quantiles $0, 0.1, \ldots, 0.9, 1$ to visualize how the \emph{distribution} of experts improves over the course of training.
Overall, we see reliable improvement of the experts with convergence at about 100 million environment transitions.

\subsubsection{Expert Performance vs. Snippet Length}
\begin{figure}[ht!]
    \centering
    \includegraphics[width=0.9\textwidth]{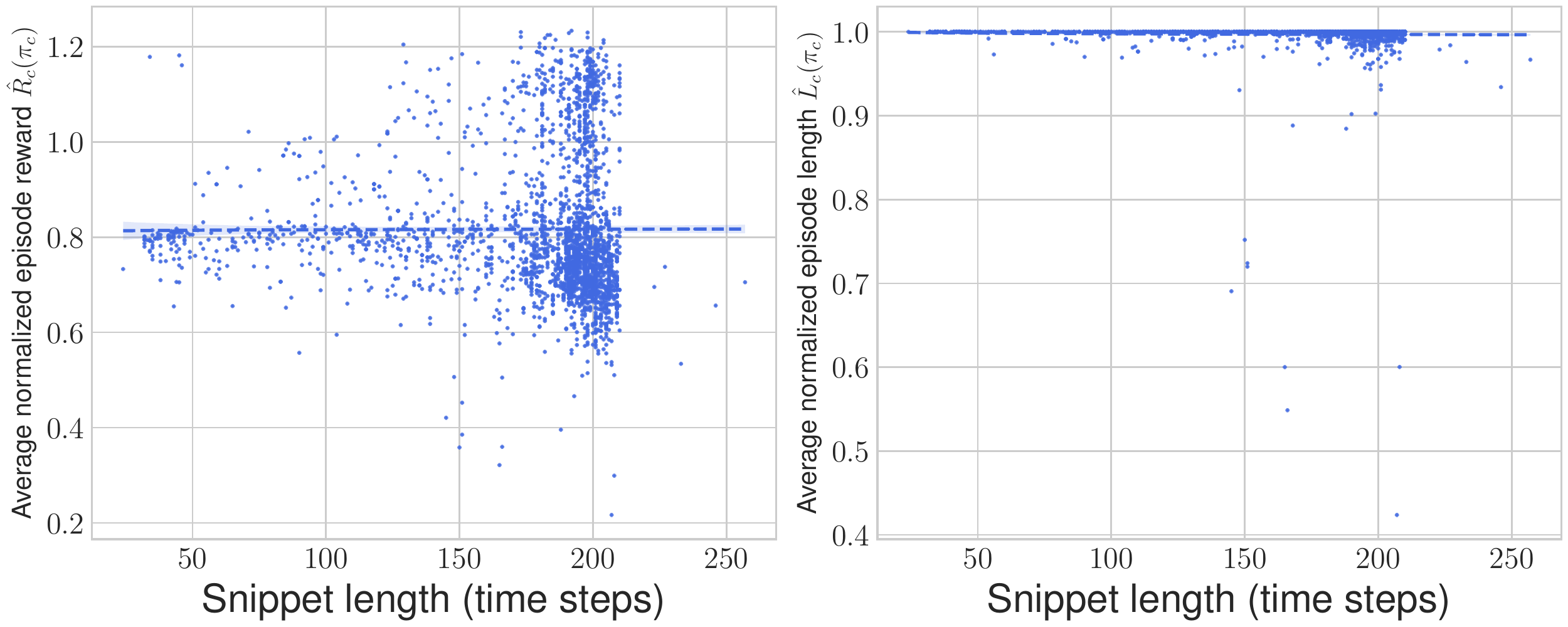}
    \caption{Scatter plot of experts' performance versus the snippet length. Here, the Gaussian noise of the experts is disabled. The performance appears to be independent of snippet length.}
    \label{fig:deterministic expert metrics vs snippet len}
\end{figure}
Here, we study whether longer snippets are ``harder'' to track by the expert.
\cref{fig:deterministic expert metrics vs snippet len} shows scatter plots of the experts' normalized episode reward and length as a function of the snippet length.
Overall, the snippet length does not appear to affect the experts' performance as indicated by the fitted curves being relatively flat.

\subsubsection{Noisy Expert Evaluations} \label{app:noisy expert evaluations}
\begin{table}[ht!]
    \centering
    \caption{Clip expert results on the \mocap snippets within \dmcontrol using the stochastic $\pi_c$.}
    \begin{tabular}{r|c|c|c|c|c}
                              & {\small Mean}    & {\small Standard deviation} & {\small Median} & {\small Minimum} & {\small Maximum} \\ \hline
    {\small Average normalized episode reward} & $0.689$ & $0.092$            & $0.690$ & $0.179$ & $0.876$ \\
    {\small Average normalized episode length} & $0.984$ & $0.029$            & $0.990$ & $0.403$ & $1.000$ \\
    \end{tabular}
    \label{tab:noisy clip expert metrics}
\end{table}

\begin{figure}[ht!]
    \centering
    \begin{subfigure}[b]{0.66\textwidth}
        \centering
        \includegraphics[width=\textwidth]{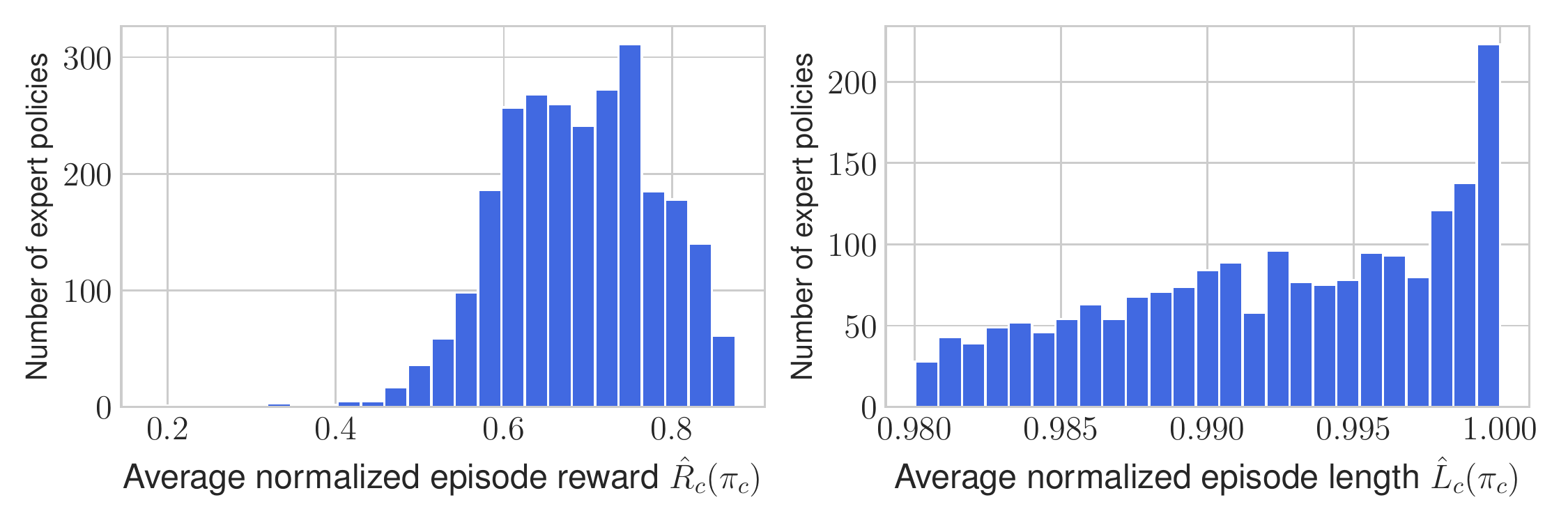}
        \caption{Episode rewards and lengths of the noisy experts.\\~}
        \label{fig:noisy clip expert metrics}
    \end{subfigure}
    \hfill
    \begin{subfigure}[b]{0.33\textwidth}
        \centering
        \includegraphics[width=\textwidth]{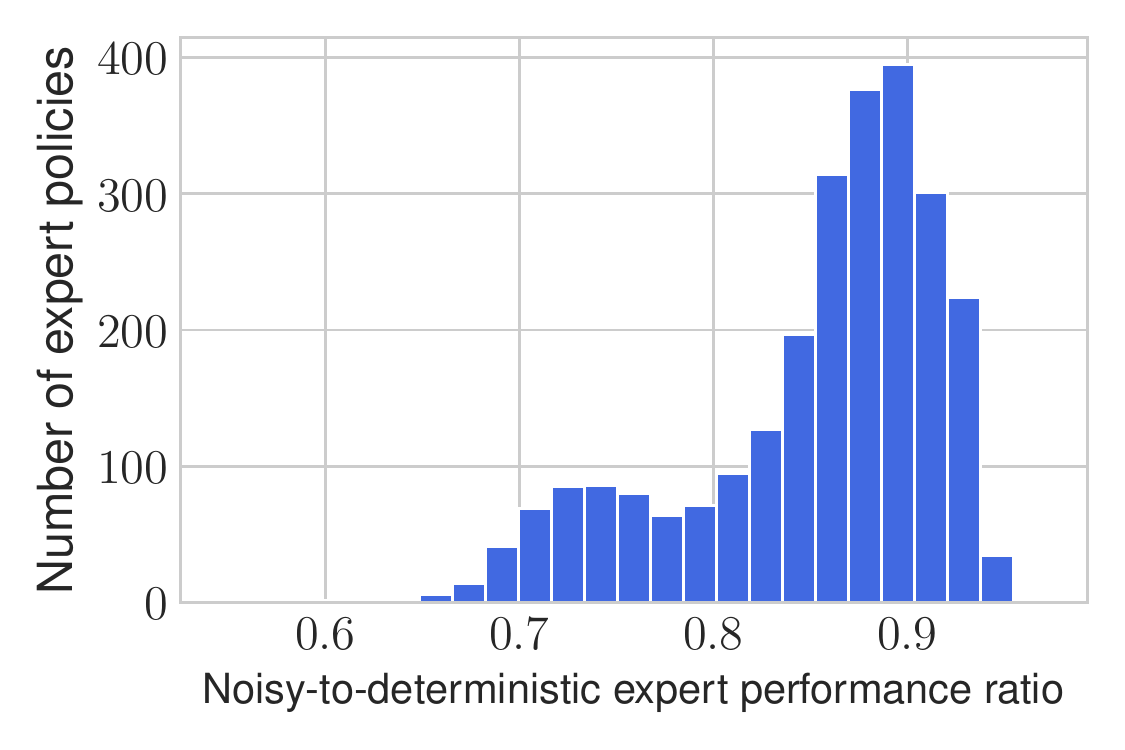}
        \caption{Performance ratio of noisy expert to deterministic expert.}
        \label{fig:noisy clip expert performance ratio}
    \end{subfigure}
    \caption{Noisy expert results on the \mocap snippets within \dmcontrol. The noisy experts incur a small performance drop from their deterministic counterparts.}
\end{figure}

Because the \dataset dataset is formed from noisy rollouts of the experts, it is sensible to assess the performance of the experts when rolled out with noise.
\cref{tab:noisy clip expert metrics} and \cref{fig:noisy clip expert metrics} show that the experts still have strong performance.
We point out the noisy experts on average attain $85\%$ of the performance of the deterministic experts~(\cref{fig:noisy clip expert performance ratio}).

\begin{figure}[ht!]
    \centering
    \includegraphics[width=0.9\textwidth]{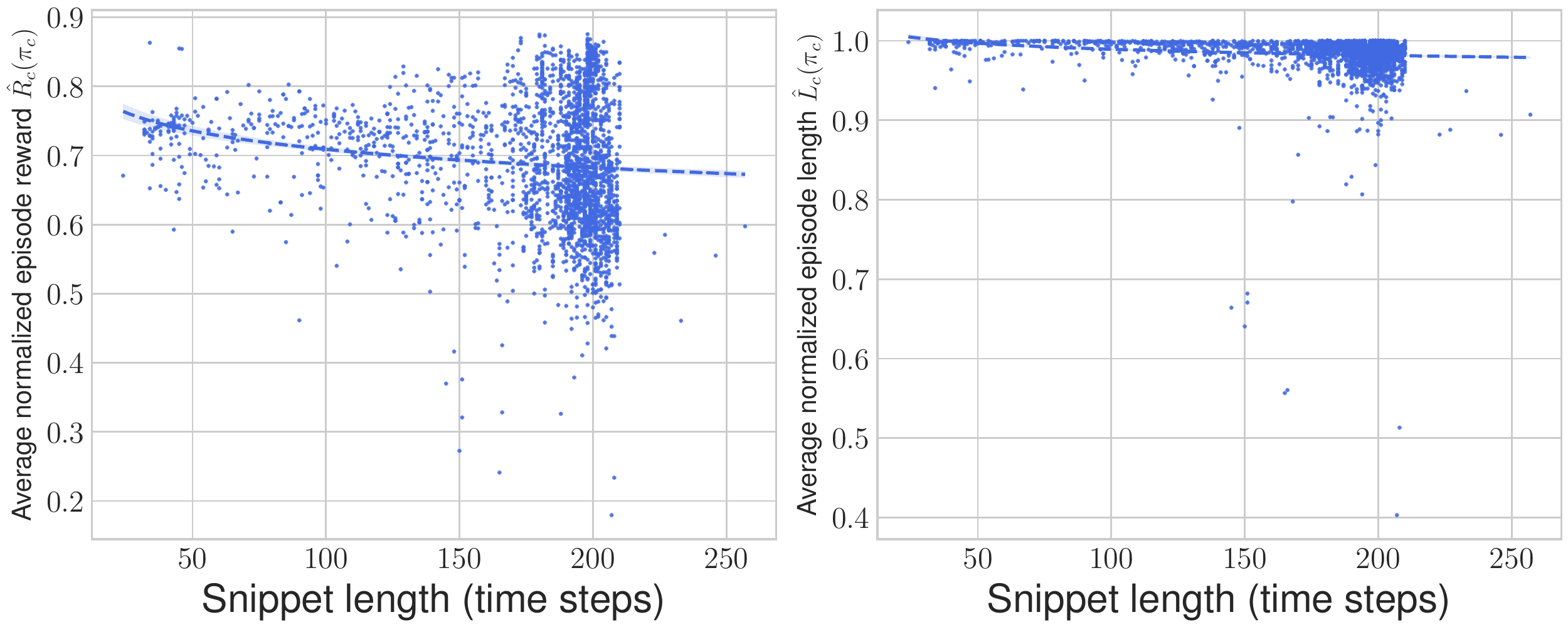}
    \caption{Scatter plot of noisy experts' performance versus the snippet length. There is a minor decrease in performance as the snippet length increases.}
    \label{fig:noisy expert metrics vs snippet len}
\end{figure}

From the scatter plot of the noisy experts~(\cref{fig:noisy expert metrics vs snippet len}), we see a minor decrease in reward and episode length as the snippet gets longer.
This is probably due to longer snippets giving more time steps for the noise to destabilize the humanoid.

\newpage
\subsection{Multi-Clip Tracking Policy}
\subsubsection{Training Curves} \label{sec:multi-clip training curves}
\begin{figure}[ht!]
    \centering
    \includegraphics[width=0.7\textwidth]{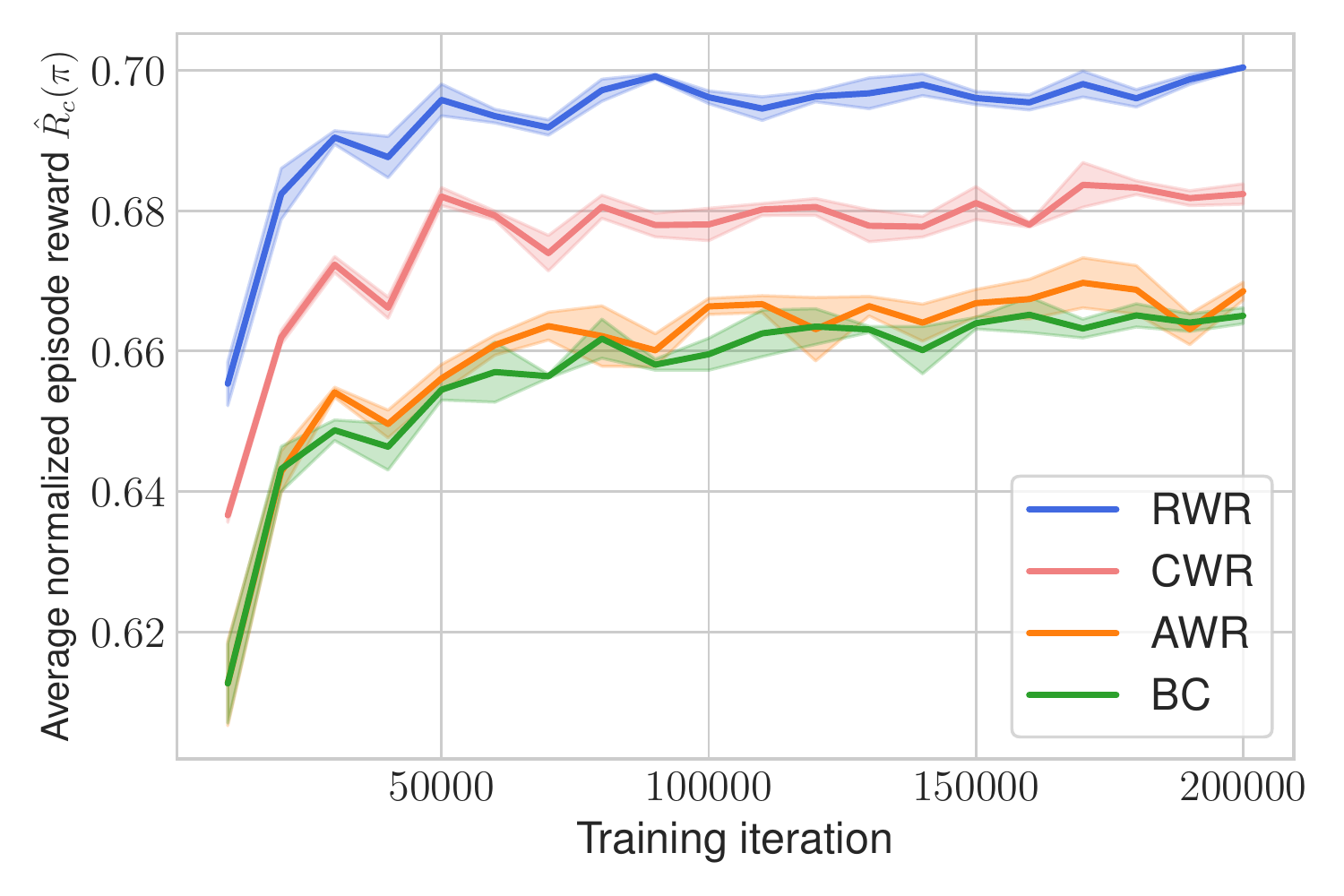}
    \caption{Multi-clip policy training curves on \mocap snippets.}
    \label{fig:multi-clip training curves}
\end{figure}

\cref{fig:multi-clip training curves} shows the reward curves for the four weighting schemes.
Overall, the reward plateaus after about \num{50000} iterations for each scheme, and reward-weighted regression performs markedly better than the other three schemes.

\subsubsection{Autoregressive Parameter $\alpha$}
\begin{figure}[h!]
    \centering
    \includegraphics[width=0.7\textwidth]{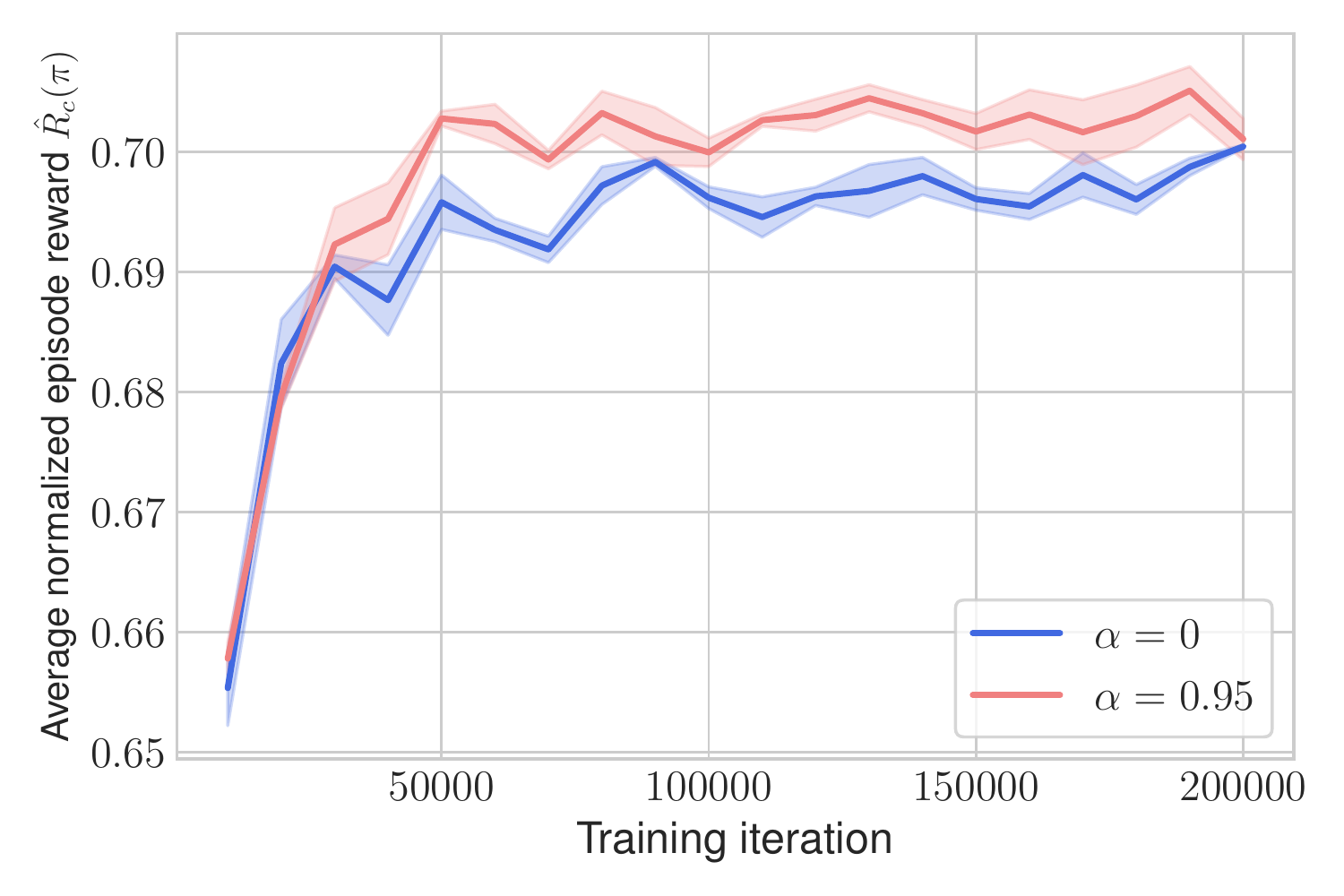}
    \caption{Comparison of multi-clip policy's performance when varying the autoregressive parameter $\alpha$ for the prior distribution $p(z_t | z_{t-1})$. Here, we use the RWR-weighting scheme. Performance is broadly similar for both values of $\alpha$.}
    \label{fig:multi-clip policy reward vs correlation}
\end{figure}

\citet{merel2019neural} found that using an autoregressive parameter of $\alpha = 0.95$ gave $50\%$ improvement in policy performance over $\alpha = 0$.
Interestingly, in our experiments we found that the performance gap is much smaller~(\cref{fig:multi-clip policy reward vs correlation}), with $\alpha = 0.95$ only giving $3\%$ improvement.
Accordingly, we set ${\alpha = 0}$ for our experiments (corresponding to a temporally independent prior of ${p(z_t | z_{t-1}) = \mathcal{N}(z_t; 0, I)}$) so that we could better control the size of the intentions $z_t$ generated by our reference encoder.

\subsubsection{Scatter Plots on Snippets and Clips}

\begin{figure}[ht!]
    \centering
    \includegraphics[width=0.9\textwidth]{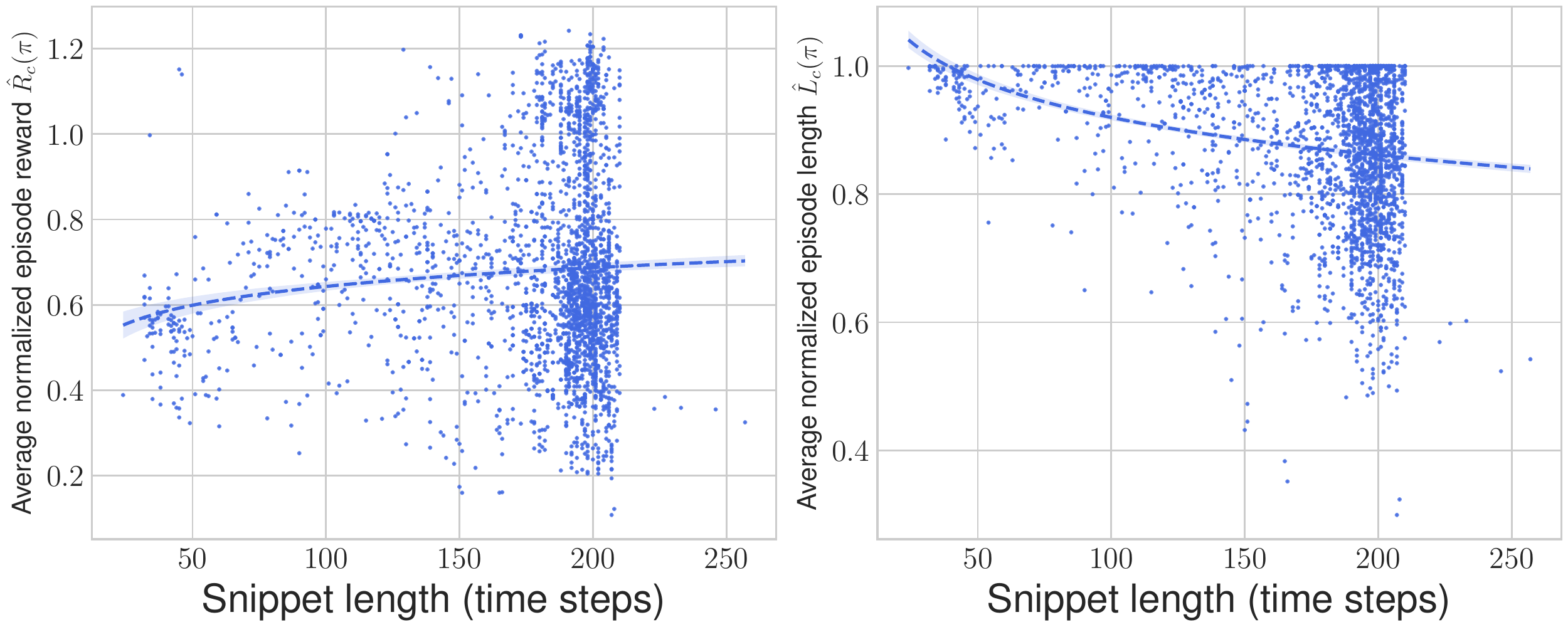}
    \caption{Scatter plot of the multi-clip policy's performance versus the snippet length. Here, the Gaussian noise of the policy is disabled. Longer snippets tend to result in lower episode lengths.}
    \label{fig:multi-clip policy metrics vs snippet len}
\end{figure}

\cref{fig:multi-clip policy metrics vs snippet len} shows the scatter plot of the multi-clip policy on all of the \mocap snippets.
Compared to the noisy experts~(\cref{app:noisy expert evaluations}), we see a more noticeable decline in episode length on long snippets.
Intuitively, this is because longer snippets allow for more opportunities for the multi-clip policy to make an episode-ending mistake.
The normalized reward, on the other hand, does not give any meaningful trends.

\begin{figure}[ht!]
    \centering
    \includegraphics[width=0.9\textwidth]{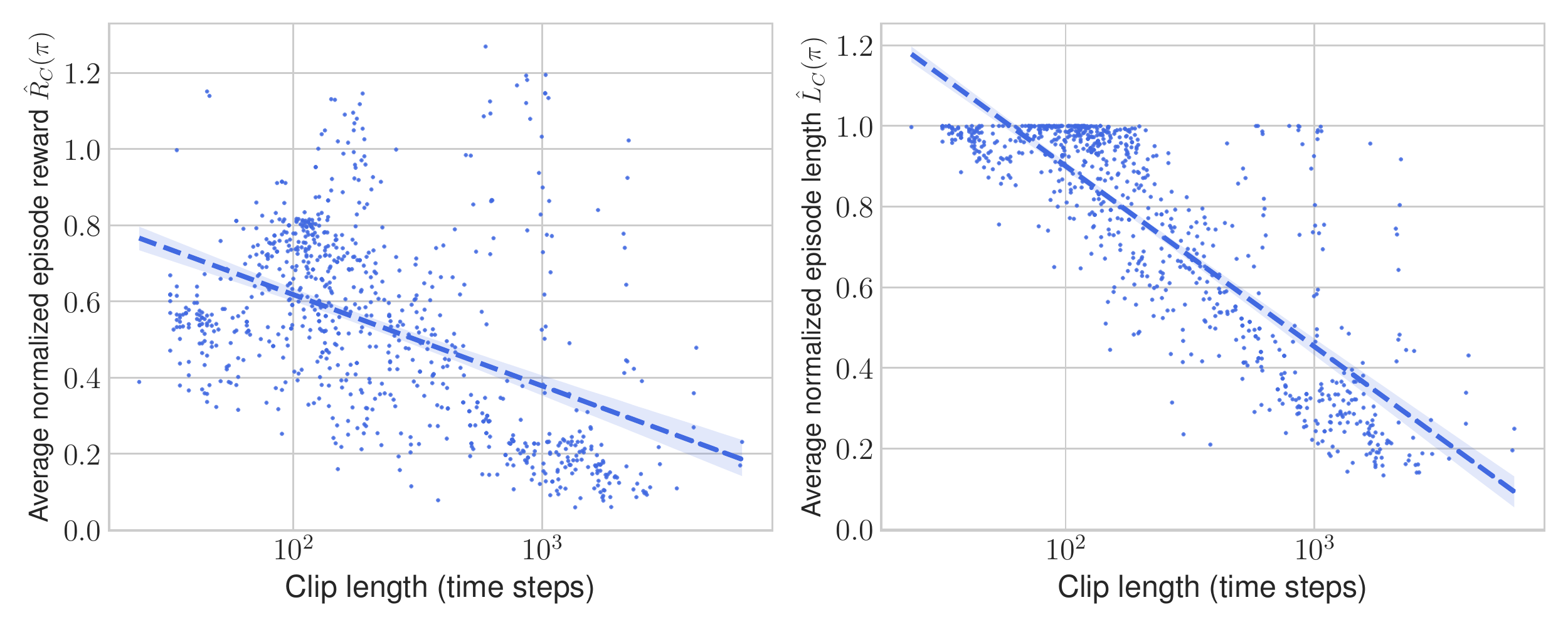}
    \caption{Scatter plot of the multi-clip policy's performance versus the clip length. Here, the Gaussian noise of the low-level policy is disabled. Longer clips tend to result in lower episode rewards and lengths.}
    \label{fig:multi-clip policy metrics vs clip len}
\end{figure}

One appealing feature of the multi-clip policy is the ability to roll out the policy on entire clips.
This also allows us to discover whether the multi-clip policy has learned to ``stitch'' together the overlapping snippets from the dataset.
\cref{fig:multi-clip policy metrics vs clip len} shows that while there are long clips that the policy can reliably track, the overall trend is that longer clips result in lower reward and episode length.
Intuitively, many clips in the \mocap dataset correspond to locomotion behaviors, which gives many opportunities for the multi-clip policy to make episode-terminating mistakes.
Usually, these mistakes correspond to the humanoid legs colliding or one of the feet making bad contact with the ground, both of which cause the humanoid to fall over.
The fragility on longer clips points to a shortcoming of \dataset: the rollouts only cover (at most) a 6-second window.
Because of this, the multi-clip policy is not trained on states that would be encountered deep into a rollout (e.g., 30 seconds into a rollout), which limits the multi-clip policy's performance on many longer clips.
Long clips that have high rewards and episode lengths usually have the humanoid standing for long periods of time while doing various arm motions.
Here, the motions are much simpler since the humanoid merely needs to maintain balance while standing still.

\end{appendix}

\end{document}